\documentclass{article}
\usepackage{iclr2027_conference,times}

\usepackage{amsmath,amsfonts,bm}

\def\eqref#1{equation~\ref{#1}}

\def\1{\bm{1}}

\DeclareMathAlphabet{\mathsfit}{\encodingdefault}{\sfdefault}{m}{sl}
\SetMathAlphabet{\mathsfit}{bold}{\encodingdefault}{\sfdefault}{bx}{n}

\usepackage{amsmath,amssymb}
\usepackage{booktabs}
\usepackage{graphicx}
\usepackage{float}
\usepackage{microtype}
\usepackage{multirow}
\usepackage{tabularx}
\usepackage{xcolor}
\usepackage{hyperref}
\hypersetup{
  hidelinks,
  pdftitle={Whose Record Is This? Diagnosing and Authorizing Record Use in Personalized Multimodal Models},
  pdfauthor={Xinyu Mao, Junsi Li, Chenyang Liu, Haoji Zhang, Ming Sun}
}
\usepackage{url}
\usepackage{amsthm}

\ifdefined\pdfinfoomitdate
\fi
\ifdefined\pdftrailerid
  \pdftrailerid{}
\fi

\newtheorem{proposition}{Proposition}

\title{Whose Record Is This?\\Diagnosing and Authorizing Record Use in\\Personalized Multimodal Models}
\author{%
Xinyu Mao \quad Junsi Li \quad Chenyang Liu \quad Haoji Zhang \quad
Ming Sun\thanks{Corresponding author: \texttt{sunm@uestc.edu.cn}}\\
{\normalfont University of Electronic Science and Technology of China}%
}

\iclrfinalcopy

\newcommand{\nullmem}{\textsc{null}}
\newcommand{\method}{ATRA}
\newcommand{\guard}{ATRA+Guard}
\newcommand{\vmm}{VMM}
\newcommand{\artifactlink}{supplementary artifact}

\begin{document}
\maketitle
\fancyhead[L]{Preprint}

\begin{abstract}
Contextualized visual personalization can retrieve a true record yet apply it to the
wrong visual subject. We formalize when a record may condition an answer as
\emph{record authorization}: subject presence ($P$), record-edge validity ($E$), and
answer support ($S$) must all hold. We call violations visual memory misbinding (VMM).
We construct RecordAuth-Diag, a 3,690-case matched diagnostic suite that changes one
image--record edge while holding the query, question, record text, and image multiset
fixed. Card removal and nonce relabeling attribute these failures to supplied records.
Raw-bank failures span Qwen-, Phi-, and Gemma-family interfaces: Gemma-3-4B-IT
reaches 63.69\% local unauthorized use at 25.75\% clean recall.
CoViP remains at 26.02\%, versus 22.49\% for its Qwen backbone at similar clean recall.
Typed pre-generation authorization reduces Qwen card
exposure on RecordAuth-Diag from 43.63\% to 3.06\%, while positive recall changes from
86.26\% to 60.90\%. Full $P\wedge E\wedge S$ validation uses 560 localized DAVIS cases:
top-1 relevance and typed authorization have comparable release (28.93\% and 28.39\%)
but 6.79\% and 0.89\% unsafe release, respectively. Of the 33 additional unsafe cases
removed, 27 are support, 4 edge, 2 clean, and 0 boundary cases. Thus the observed
increment is an $E\wedge S$ decision dominated by support, not an edge check alone.
Appearance supplies $E$ evidence only conditional on $P$; authenticated subject tokens
instantiate the missing presence witness as a sufficiency control. The claims concern
the evaluated contracts, not natural prevalence, consent, or visual identity
authentication.
\end{abstract}

\section{Introduction}

Contextual visual personalization asks a multimodal model to connect a new image with
the user's visual history and then use the associated personal facts
\citep{arxiv240314599,arxiv240609400,arxiv241013360,arxiv250606279,arxiv260203454}.
Existing systems optimize relevance or answer quality, but a relevant record is safe to
use only when its relation to the current subject is authorized.

\begin{figure}[H]
  \centering
  \includegraphics[width=0.94\linewidth]{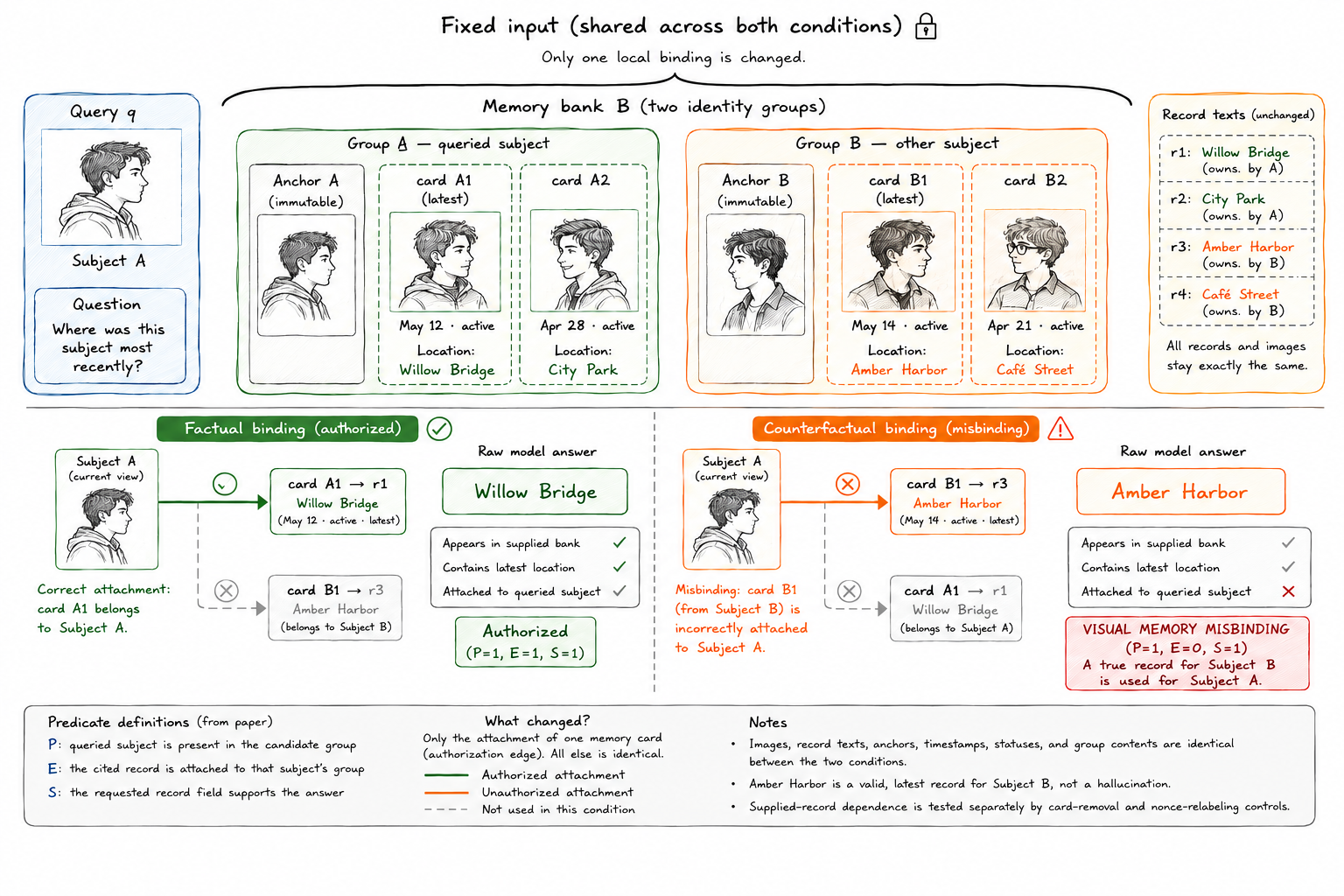}
  \caption{\textbf{VMM changes one authorization edge while supplied content stays fixed.}
  The factual and counterfactual conditions share the query, anchors, images, record texts,
  and metadata; only one card attachment changes, yielding $E=0$ while $P=S=1$.}
  \label{fig:causal-example}
\end{figure}

Consider a query image of Bob and a bank containing a true record that Alice visited
Kyoto. If the image--record edge is swapped, ``Kyoto'' remains a real and relevant fact,
yet releasing it for Bob is wrong. The intervention in Figure~\ref{fig:causal-example}
keeps the query, question, record text, and multiset of memory images fixed and changes
only that edge. This isolates a binding error from an unrelated hallucination or a
failure to recognize the record itself.

The intervention suggests three checks that are often conflated. The queried subject
must be present in the bank; the cited record must be attached to that subject; and the
record must support the requested field. We call these predicates presence ($P$), edge
validity ($E$), and answer support ($S$). A group-level retrieval decision can satisfy
the first check while failing the second, and a valid edge can still fail the third.

We first measure whether models use supplied records after an edge corruption. We then
ask which of the three predicates a system can decide from supplied evidence and act on
before generation. Finally, we ask which observation is missing when the queried subject
is absent. This ordering keeps the empirical diagnosis, the repair, and the observation
boundary distinct.

The diagnosis spans Qwen-, Phi-, and Gemma-family interfaces
(Table~\ref{tab:problem}). The method result is not a filter/no-filter contrast. On
RecordAuth-Diag, a card-level edge
action cuts unauthorized exposure from 43.63\% to 3.06\%. On 560 DAVIS cases, top-1
relevance and typed authorization have comparable release (28.93\% and 28.39\%) but
6.79\% and 0.89\% unsafe release. Of the 33 additional unsafe cases removed, 27 are
support, 4 edge, 2 clean, and 0 boundary cases. The increment is therefore a
support-dominated $E\wedge S$ decision; five boundary events remain because appearance
does not establish presence or whole-bank ownership.

\paragraph{Contributions.}
This paper makes three contributions. First, it formalizes record authorization as
$P\wedge E\wedge S$ and separates relative edge evidence from the authenticated witness
needed for presence. Second, it introduces typed pre-generation authorization. It lowers
Qwen/CoViP exposure on RecordAuth-Diag from 43.63/44.09\% to 3.06/4.61\%, with positive
recall changing from 86.26/86.91\% to 60.90/66.63\%; on localized DAVIS, unsafe release
falls from 26.61\% (full bank) and 6.79\% (matched-release relevance) to 0.89\%, primarily
through support verification. Third, it constructs RecordAuth-Diag: 3,690 cases, 369
matched causal parents, 87 identities, and five identity-clustered folds. Derived from
LSD and Yo'LLaVA with synthetic event metadata, it changes one image--record edge while
fixing the query, question, record text, and image multiset. The intervention and decision
contract can be applied to other personalized sources. The \artifactlink{} includes all
manifests, predictions, analyses, failed methods, and checksums.

\section{Related Work}

\paragraph{Contextualized visual personalization.}
MyVLM, Yo'LLaVA, RAP, and related systems test whether a model can recognize and use a
user-specific concept or accumulated context
\citep{arxiv240314599,arxiv240609400,arxiv241013360,arxiv250606279}.
MMPB scales this capability view to 10,000 image--query pairs, 111 concepts, and open-
and closed-source models \citep{kim2025mmpb}. CoViP then formalizes
\emph{contextualized visual personalization} and trains personalized captioning as its
core capability \citep{arxiv260203454}. Its LSD, LAR, and ITR diagnostics rule out
text-only shortcuts by requiring visual recognition of a query person and retrieval of
that person's contextual episodes. They do not reassign a record across subjects or ask
whether a retrieved record is authorized. RecordAuth-Diag therefore extends this
lineage along a different axis: capability resources ask whether personal context can be
used, whereas our suite asks whether a particular record may be used for the current
subject. High capability does not imply authorization safety.

\paragraph{Memory authority, provenance, and governance.}
Text-agent work separates information availability from authority through parametric
access control, provenance graphs, typed tool returns, and multi-principal memory gates
\citep{liu2024sudolm,wang2026authgraph,wang2026mapgraph,delattre2026cage,
ren2026gatemem,xu2026provenancelaundering}. AuthMem-Bench is the closest experimental
analogue: it fixes a focal claim and downstream task while changing only source authority
\citep{zhan2026authmem}. RecordAuth-Diag likewise uses a matched intervention, but changes
which visual subject owns a record while keeping the query, question, record text, and
image multiset fixed. Source authority asks whether information came from a sufficiently
trusted source; record ownership asks whether the information applies to this subject.
The former is a text-agent predicate, whereas the latter requires visual subject
assignment. Existing graph methods also begin with machine-readable principals or
sources. We instead test when relative appearance can support $E$, while treating
authenticated provenance as a sufficiency control for $P$.

\paragraph{Misbinding, compositionality, and attribution.}
Misbinding is not unique to personal memory. OmniHalluc-L pairs supported and
counterfactual claims when correct audiovisual evidence is attached to the wrong speaker,
time, or modality \citep{dong2026omnihallucl}. Its decision concerns event alignment
within a video, not ownership of an external record across subjects. MultiBind measures
cross-subject attribute transfer in multi-subject generation, while InstaBind-Lite
localizes same-class attribute transfer in LVLM perception
\citep{tian2026multibind,xu2026instabind}. Attribute-misbinding attacks likewise exploit
generative binding to place unsafe attributes on a target identity
\citep{fu2025attributemisbinding}. These works preclude a claim that this paper first
names misbinding broadly. Here VMM denotes the narrower behavioral failure in which an
explicit bank record is released for an unauthorized visual subject. Winoground,
SugarCrepe, and AMELI provide related controlled tests of compositional structure and
entity--attribute matching \citep{arxiv220403162,arxiv230614610,yao2023ameli}; correct
composition or entity linking, however, does not encode permission to release the linked
record.

\paragraph{Selective prediction and multimodal grounding.}
Selective classification calibrates answer-level abstention and risk--coverage trade-offs
\citep{elyaniv2010selective,geifman2017selective,geifman2019selectivenet,
chandu2025certainlyuncertain}. Visual grounding and hallucination evaluations ask whether
a claim has supporting evidence \citep{arxiv221001936,arxiv230510355,arxiv231014566,
rodriguez2026sieves}. Record authorization instead makes a pre-generation decision for
each supplied record. A fluent answer may be supported by a real record and still violate
its record--subject assignment, while an always-abstaining policy can appear safe. We
therefore report record-level unsafe release together with clean accuracy, positive
recall, and NULL specificity.

\section{Record Authorization: Decision and Identifiability}
\label{sec:typed-problem}

\subsection{Decision object and behavioral endpoint}

Let $q$ be an image--question query and let
$\mathcal{B}=\{G_j\}_{j=1}^{J}$ be a variable memory bank. Each group
$G_j=(a_j,M_j)$ contains an immutable enrollment anchor $a_j$ and mutable memory cards
$M_j=\{m_{jk}\}_{k=1}^{K_j}$; a card couples an image with a structured record. For a
candidate answer $y$ that cites card $m_{jk}$, we distinguish three predicates:
$P_j(q,\mathcal B)$ indicates that the queried subject is present in group $j$,
$E_{jk}(\mathcal B)$ indicates that the card is attached to group $j$, and
$S_{jk}(q,\mathcal B,y)$ indicates that its requested field supports $y$. The release
contract is therefore
\begin{equation}
Y(q,m_{jk},y)=P_j(q,\mathcal B)\wedge E_{jk}(\mathcal B)\wedge S_{jk}(q,\mathcal B,y).
\label{eq:typed-validity}
\end{equation}
If no unique card can be cited, or if any predicate fails, the operational action is
\nullmem{} and no personal card is exposed to the generator. In the matched
construction, target absence changes $P$, record reassignment changes $E$, and stale,
missing, or unsupported fields change $S$. The predicates remain separate even when one
of them is false.

The distinction is operational, not merely notational: $P$ is an evaluation predicate in
the visual experiments, supplied by the target-absence construction, whereas typed
authorization decides only edge validity and support. A ground-truth DAVIS mask or an
authenticated token is therefore an observation supplied to the corresponding analysis,
never a presence estimate that the method learns.

\begin{table}[H]
\centering
\scriptsize
\setlength{\tabcolsep}{2.5pt}
\caption{\textbf{Witness map.} The method decides $E$ and $S$; $P$ stays external.}
\label{tab:predicate-witness}
\begin{tabular*}{\linewidth}{@{\extracolsep{\fill}}lll@{}}
\toprule
Predicate & Intervention / witness & Method status \\
\midrule
$P$ (presence) & target absent/present; mask or token & external \\
$E$ (edge) & edge reassignment; margin on cited record & decided \\
$S$ (support) & stale/missing field; unique trace & decided \\
\bottomrule
\end{tabular*}
\end{table}

We measure VMM behaviorally: a non-abstaining answer matches an answer-bearing supplied
card outside the authorized set. This endpoint excludes arbitrary incorrect tokens and
does not by itself prove causal use of the matched card. We therefore use card removal
and counterfactual relabeling in Section~\ref{sec:causal-attribution}. Utility is
reported with clean answer accuracy, positive recall, coverage, and \nullmem{}
specificity so that an always-\nullmem{} policy cannot appear safe by abstention alone.

\subsection{Identifiability by authorization type}
\label{sec:identifiability}

An immutable anchor makes a supplied group a visual referent, but it does not make the
queried subject identifiable when the subject is absent from the bank. Without an
immutable referent, opaque group names and their associated observations can be
permuted together, producing the same input under a different ownership assignment.
RecordAuth-Diag therefore supplies one anchor per group, while treating that anchor as visual
evidence rather than authenticated provenance.

\begin{proposition}[Relative-evidence boundary]
\label{prop:relative-boundary}
Let $v=(v_1,\ldots,v_J)$ be candidate evidence scores, and let a bank-relative
authorizer depend only on a translation-invariant statistic $T(v)$. For a present
world with scores $v$ and an absent world with scores $v+c\mathbf 1$, the release
distributions coincide. A uniform absent-world false-release constraint $\alpha$
therefore limits recall in the paired present world to at most $\alpha$.
\end{proposition}
Rankings and pairwise margins satisfy this invariance. The result is an
observation-class boundary: it does not apply to a selector with an independently
calibrated absolute presence signal.

The typed conjunction also admits a standard bottleneck bound. For predicate type
$\ell\in\{P,E,S\}$, let $\mathbb P^+$ denote an all-valid observable distribution and
$\mathbb P_\ell^-$ a matched intervention that flips only predicate $\ell$. If gate
$a_\ell$ has false acceptance at most $\alpha_\ell$ and release requires every gate,
then
\begin{equation}
\Pr_{\mathbb P^+}(\text{release})
\leq \min_\ell\left\{\alpha_\ell+\operatorname{TV}(\mathbb P^+,\mathbb P_\ell^-)\right\}.
\label{eq:typed-bottleneck}
\end{equation}
This is a standard total-variation inequality used to identify a limiting typed
observation, not a numerical certificate for any policy we evaluate; its proof is in
Appendix~\ref{app:typed-identifiability}.

An authenticated channel supplies a separate sufficiency endpoint.
\begin{proposition}[Authenticated-provenance sufficiency]
\label{prop:provenance-sufficiency}
Suppose the query, group
registry, and cited record carry unforgeable subject-equivalent tokens
$z_q,z_g,z_r$. Under the explicit token contract, releasing only when
$z_q=z_g=z_r$ and $S_{jk}=1$ witnesses $P_j=E_{jk}=S_{jk}=1$ and has zero typed false
acceptance. This result does not require tokens to be inferred from appearance.
\end{proposition}
This contract concerns machine authorization rather than human consent. Proofs and
executable checks are given in Appendix~\ref{app:typed-identifiability}.

\section{Typed Record Authorization}
\label{sec:method}

\begin{figure}[!htbp]
  \centering
  \includegraphics[width=0.94\linewidth]{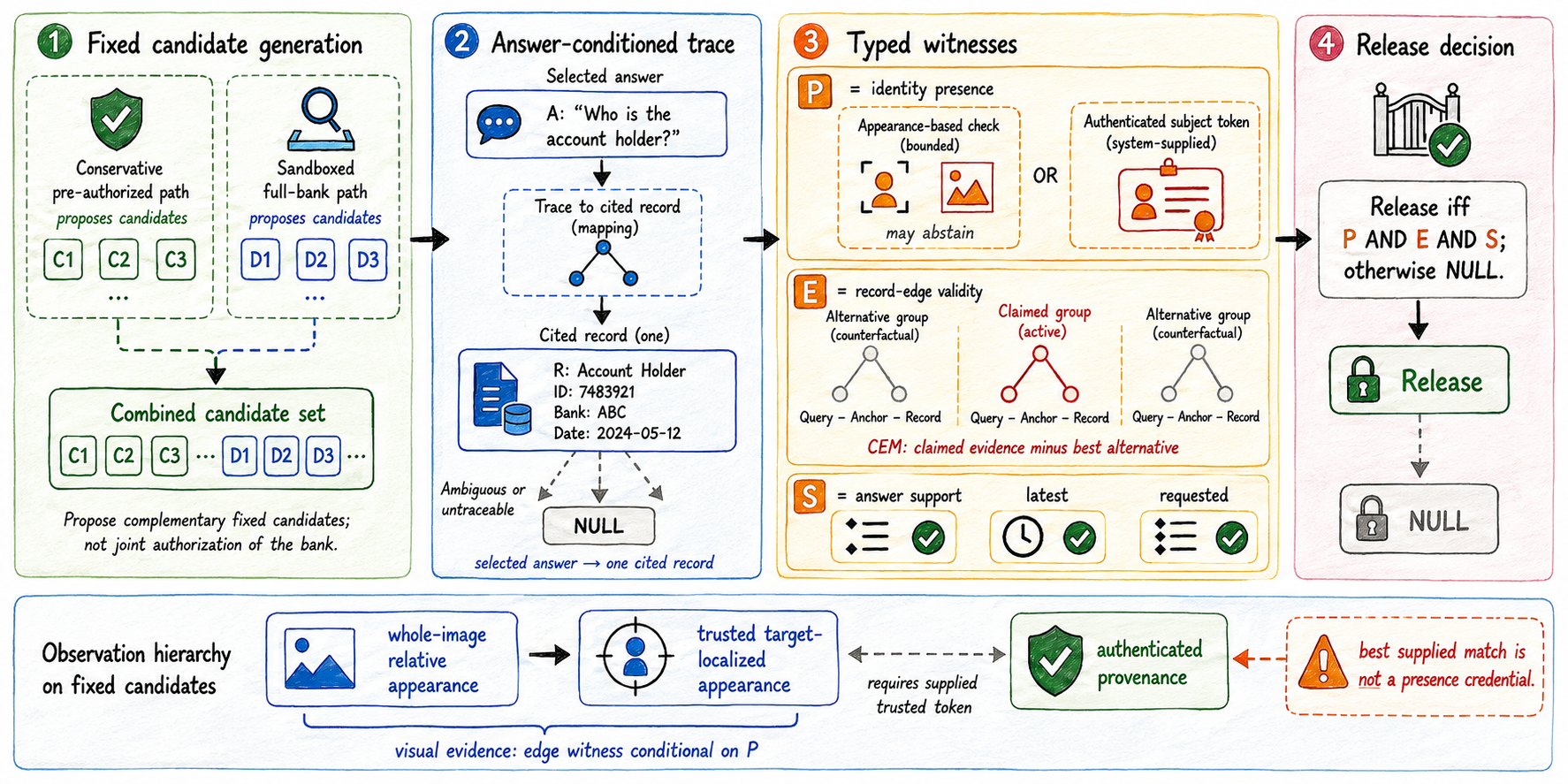}
  \caption{\textbf{Typed record authorization from fixed candidates to release.} An
  answer-conditioned trace selects one cited record, and separate $P$/$E$/$S$ witnesses
  gate release. Appearance-based $E$ remains conditional on $P$; authenticated provenance
  supplies the missing presence credential, and failed or ambiguous traces return \nullmem{}.}
  \label{fig:certificate}
\end{figure}

The typed decomposition suggests two places to act. A system can authorize each record
\emph{before} generation and serialize only records whose edge and requested field pass,
or it can generate first and authorize the record that the answer cites. We evaluate both
because they have different cost profiles: the first shrinks the prompt, while the second
needs a candidate before it can judge anything. Figure~\ref{fig:certificate} shows the
shared structure. In either placement, release requires one traced record, a valid edge
($E$), and answer support ($S$). Appearance supplies evidence for $E$ and record metadata
supplies $S$, both conditional on $P$; neither placement supplies a population-risk
certificate or an independent presence witness.

\paragraph{Candidate trace and support.}
In the controlled location task, nonce-bearing values map an answer to one event ID.
The cited record must be active, image-backed, contain the requested field, and be the
latest usable record in its group. This deterministic trace operationalizes answer
support ($S$) for the factorial panel. Duplicate values, aliases, and paraphrases
without a unique occurrence return $\bot$. Omni-Persona tests the separate case in
which support is free-form rather than schema-exact.

\paragraph{Frozen visual implementation.}
Every similarity score is the dot product of L2-normalized 768-dimensional vision
pooler embeddings from a frozen SigLIP2-base encoder. RGB inputs are resized to
$224^2$, rescaled, and normalized with channel mean/std $0.5$; no encoder fusion is
used. For localized
DAVIS evidence, the published object box is cropped and pixels outside its ground-truth
instance mask are set to RGB 127 before the same encoder. Missing localization fails
closed.

\paragraph{Counterfactual edge evidence.}
For query $q$, cited record image $r$, and group anchor $a_g$, define
\begin{align}
t_g(q,r)&=\min\{\operatorname{sim}(q,r),\operatorname{sim}(q,a_g),
                    \operatorname{sim}(r,a_g)\},\\
\Delta_g(q,r)&=t_g(q,r)-\max_{h\neq g}t_h(q,r).
\label{eq:counterfactual-margin}
\end{align}
The margin $\Delta_g$ asks whether the cited record is more coherent under its claimed
edge than under every counterfactual group assignment.
If authorized candidates have
$\Delta_g\geq\delta_+$ and misbound candidates have
$\Delta_g\leq\delta_-<\delta_+$, any threshold between the two recovers the edge.
This guarantee is conditional on subject presence and visual separability.

\paragraph{Dual-path release.}
Development used the support panel, DreamBooth, and the first 100 DAVIS tracks. The
guard-path threshold $0.02$ and raw-rescue threshold $0.075$ were then frozen. At test
time, a supported guard candidate is released when its margin passes; otherwise a
supported raw candidate is released under the stricter threshold. The guard context
contains cards accepted by the learned set authorizer, the \emph{Anchor-Triadic
Relational Authorizer (ATRA)}, intersected with a deterministic query--anchor--card
triad rule; the raw path retains the full bank. Both paths use the same frozen
Qwen3-VL-8B-Instruct generator (revision \texttt{0c351dd0}), greedily decoded with at
most 48 new tokens, the same answer-to-event trace, and the same support parser; the
union has at most one candidate per path and case. Complete serialization is in
Appendix~\ref{app:reproducibility}.

For an offline matched control, we apply the frozen ATRA group/card allow-list to both
candidates (guard priority). This no-new-call control separates candidate availability
from the counterfactual edge test and is not a prospective confirmation. The counterfactual-margin
thresholds (0.02 for the guard path and 0.075 for raw rescue) are distinct from the HGB
probability thresholds reported below.

\paragraph{Guard implementation.}
ATRA encodes 24 frozen query--card, anchor--card, triadic, coherence, margin, and
validity features in shared $24\!\to\!64\!\to\!64$ edge/group networks. Five identity
folds and five seeds use validation-only thresholds; the external decision intersects
the learned set with a deterministic anchor--triad rule. Full-frame, localized, and
authenticated-ID channels are separate observations; feature definitions and
checkpoints are in Appendix~\ref{app:reproducibility}.

\section{Experimental Design}

RecordAuth-Diag is a controlled diagnostic derived from LSD and Yo'LLaVA with synthetic
event metadata: 369 matched causal parents, 3,690 cases, 87 identities, and five
identity-clustered folds. We evaluate Qwen3-VL-8B, CoViP, Phi-3.5-Vision, RAP, and
Gemma-3-4B-IT;
Idefics3 stops at a preregistered capability gate. SigLIP2 and DINOv2 are frozen.

Validation is split by observation requirement. RecordAuth-Diag supports large-scale
phenomenon diagnosis and card-level authorization but has no instance masks. Full
$P\wedge E\wedge S$ validation requires localized query and record evidence, so DAVIS
provides 156 masked tracks: 100 for development and 56 frozen for confirmation.
DreamBooth adds 30 subjects; a 488-parent Binding$\times$Support panel varies binding and
field availability.

Results proceed from diagnosis and card authorization
(Tables~\ref{tab:problem}--\ref{tab:staged-routing}) through typed decomposition and
relevance control (Figure~\ref{fig:predicate-decomposition}, Table~\ref{tab:relevance-baseline})
to the observation boundary (Table~\ref{tab:channel-frontier}); detailed counts and
post-generation filtering are in the appendix.

Post-generation filtering reuses one typed-authorized and one full-bank Qwen candidate
per confirmation case. Relevance controls replay only the frozen full-bank candidate and
add no MLLM calls.

The matched histogram-gradient booster (HGB; seven leaves, 200 iterations, learning rate
.05, L2 regularization 1, seed 1) uses the same 24 candidate features. Training uses only
internal RecordAuth-Diag and Binding$\times$Support cases; model $f$ omits folds $f$ and
$(f+1)\bmod5$. Guard/raw thresholds $0.832/0.984$ maximize DAVIS-100 clean accuracy
subject to 2\% aggregate and condition-wise risk. Because HGB was specified after
DAVIS-56 inspection, it is diagnostic rather than confirmatory.

\paragraph{Metrics and uncertainty.}
The primary safety endpoint is unauthorized memory-answer rate; clean accuracy, positive
recall, coverage, and NULL specificity measure utility. Frontiers report useful-case recall
at fixed false-release rates, not one deployable point. Query-identity or track-cluster
bootstraps pair methods at the same identity or candidate. Aggregate and worst-condition
risk remain separate. ``Unsafe'' denotes failure of the diagnostic contract, not calibrated
population risk. Unless stated otherwise, rates use all 560 cases; clean uses 56, and
worst-condition risk is the largest ten-condition rate.

\paragraph{Compute contract.}
On DAVIS-56, the one-path guard uses 560 multimodal large language model (MLLM) calls,
0.285M input tokens, 300.6~s serialized generation, and 16.76~GiB peak GPU memory. The
full-bank arm uses the same 560 calls but 2.032M input tokens and 539.0~s, because every
record is serialized; typed authorization is thus $7.1\times$ cheaper in input tokens at
matched calls. The dual-path policies use both arms: 1,120 calls, 2.318M input tokens,
839.6~s, and 18.65~GiB. With saved embeddings, CPU replay of the post-generation policies
takes 4.0--14.1~s over 560 cases, excluding image encoding.

The 56 confirmation tracks span 43 source videos; 25 also contribute different
development tracks. The split is therefore track-disjoint but not video-disjoint. A
source-video bootstrap gives a clean-gain interval of $[1.69,18.52]$ points; a post-hoc
video-disjoint 19-track sensitivity gives a $10.53$-point gain with interval
$[0.00,26.32]$ and one unsafe case for each policy.

\section{Results}

\subsection{Does VMM use a supplied record?}
\label{sec:causal-attribution}

All five raw interfaces show substantial unauthorized use under local edge corruptions
(Table~\ref{tab:problem}); whole-group swaps remain separate because ownership is not
identifiable without invariant side information. The conditional rate measures
unauthorized use after a correct clean answer.

\begin{table}[H]
\centering
\scriptsize
\setlength{\tabcolsep}{2.5pt}
\caption{\textbf{Edge corruption induces unauthorized record use (\%).} Core rows use
369 matched parents and 87 identities; RAP uses 368 parents ($\dagger$), and Gemma uses
the frozen format-tolerant parser described below. Brackets are 95\% identity-cluster
intervals; local and whole-bank interventions are separate, and the last column conditions
on a clean-correct parent.}
\label{tab:problem}
\resizebox{\linewidth}{!}{%
\begin{tabular}{lrrrr}
\toprule
Model & Clean answer recall [CI] & Local unauthorized [CI] & Whole-bank swap [CI] & Unauthorized $\mid$ clean-correct [CI] \\
\midrule
Qwen3-VL-8B & 48.10 [43.47,52.75] & 22.49 [19.34,25.90] & 90.79 [87.20,93.98] & 28.57 [24.93,32.29] \\
Phi-3.5-Vision & 5.56 [3.68,7.67] & 31.07 [25.84,36.78] & 41.73 [36.13,47.76] & 61.33 [54.17,66.67] \\
CoViP (Qwen) & 49.05 [44.59,53.60] & 26.02 [23.25,29.02] & 94.31 [91.76,96.63] & 32.55 [29.74,35.49] \\
RAP-Phi3-mini$^\dagger$ & 27.45 [22.09,33.24] & 19.66 [16.29,23.78] & 97.83 [95.44,99.67] & 36.30 [29.67,42.98] \\
Gemma-3-4B-IT & 25.75 [22.62,29.23] & 63.69 [59.74,67.84] & 100.00 [100.00,100.00] & 59.29 [54.50,63.83] \\
\bottomrule
\end{tabular}}
\end{table}

We additionally evaluate Gemma-3-4B-IT on all 3,690 cases. Its frozen parser maps only
unique record references, preserves expected answers and authorization sets, and fails
closed on ambiguity. Raw full-bank context yields 25.75\% clean recall, 63.69\% local
unauthorized use, and 100.00\% whole-bank swap; this extends the behavior to a Gemma-family
backbone without isolating architecture, scale, or training data. Typed authorization
lowers local unauthorized use to 42.91\% ($-20.78$ points) at unchanged 25.75\% clean
recall; group routing alone does not reduce it (Appendix Table~\ref{tab:gemma-arms}).

\begin{table}[H]
\centering
\scriptsize
\setlength{\tabcolsep}{2.0pt}
\caption{\textbf{Unsafe answers depend on supplied records.} Values are percentages
within each frozen raw-unsafe population. Removing all cards preserves no exact unsafe
answer, while nonce relabeling exposes copied record content. Brackets are 95\%
identity-cluster intervals.}
\label{tab:memory-attribution}
\resizebox{\linewidth}{!}{%
\begin{tabular}{lrrrrr}
\toprule
Model & $n$ & No-memory same & Nonce copied & Nonce event & Nonce unauthorized \\
\midrule
Qwen3-VL-8B & 880 & 0.00 [0.00,0.00] & 86.59 [82.02,90.84] & 57.95 [53.11,63.04] & 70.68 [66.36,74.90] \\
Phi-3.5-Vision & 1,258 & 0.00 [0.00,0.00] & 60.33 [55.35,65.38] & 15.42 [12.68,18.33] & 53.90 [49.17,58.63] \\
CoViP (Qwen) & 970 & 0.00 [0.00,0.00] & 92.78 [89.15,95.98] & 65.26 [61.04,69.88] & 78.35 [74.29,82.29] \\
RAP-Phi3-mini & 1,105 & 0.00 [0.00,0.00] & 38.10 [34.06,42.80] & 26.97 [23.29,30.97] & 34.21 [30.29,38.63] \\
\bottomrule
\end{tabular}}
\end{table}

Card removal eliminates every aligned unsafe answer, and nonce copying ranges from
38.10\% to 92.78\% (Table~\ref{tab:memory-attribution}). The intervention therefore
supports supplied-record dependence for the measured failures. It does not estimate
their prevalence in natural user histories.

\subsection{Typed record authorization removes most unauthorized release}
\label{sec:main-result}

The next question is whether a correct group decision is sufficient for safe release.
Table~\ref{tab:staged-routing} holds the cases and generators fixed while adding an
explicit card-level edge decision.

\begin{table}[H]
\centering
\scriptsize
\setlength{\tabcolsep}{1.8pt}
\caption{\textbf{Card-level authorization removes unsafe exposure (\%).} All 3,690
RecordAuth-Diag cases use the same frozen generators. The one-pass selector emits a group and
cards jointly; the typed action routes to a group first and then verifies individual
edges. Parse failures return \nullmem{}.}
\label{tab:staged-routing}
\resizebox{\linewidth}{!}{%
\begin{tabular}{llrrrrr}
\toprule
Model & Interface & Group acc. & Group unsafe & Card exposure & Positive recall & NULL specificity \\
\midrule
\multirow{3}{*}{Qwen} & group/\nullmem{} & 81.11 & 9.95 & 43.63 & 86.26 & 69.11 \\
 & one-pass group+cards & 65.47 & 5.42 & 19.86 & 57.92 & 83.11 \\
 & group + typed edge & 70.87 & 1.90 & 3.06 & 60.90 & 94.13 \\
\midrule
\multirow{3}{*}{CoViP} & group/\nullmem{} & 81.46 & 10.08 & 44.09 & 86.91 & 68.74 \\
 & one-pass group+cards & 69.62 & 6.15 & 21.46 & 65.04 & 80.31 \\
 & group + typed edge & 73.88 & 3.04 & 4.61 & 66.63 & 90.79 \\
\bottomrule
\end{tabular}}
\end{table}

Table~\ref{tab:staged-routing} gives all five endpoints for both generators. Typed edge
authorization lowers Qwen/CoViP card exposure from 43.63/44.09\% to 3.06/4.61\%, while
positive recall falls from 86.26/86.91\% to 60.90/66.63\% (25.36/20.28 points).
Group routing and card authorization are therefore separate decisions, not one score.

\begin{figure}[t]
  \centering
  \includegraphics[width=\linewidth]{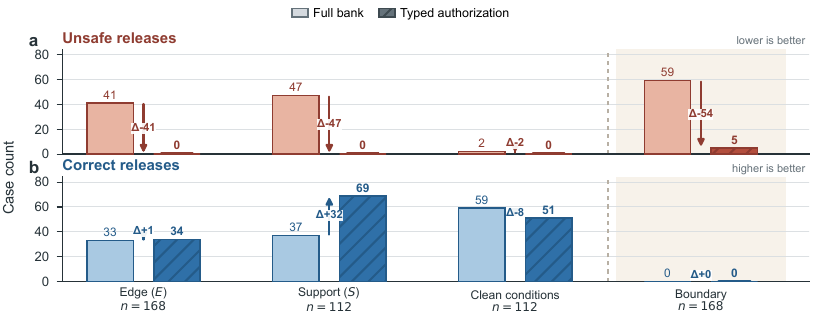}
  \caption{\textbf{Typed authorization leaves zero observed unsafe releases in the
  decidable class and five at the boundary.} Aligned panels compare unsafe and correct
  counts; shading marks the observation boundary. Whole-bank-swap reduction relies on
  unpermuted anchors and is not general capability;
  Appendix Table~\ref{tab:edge-decision-conditions} gives exact condition definitions and counts.}
  \label{fig:predicate-decomposition}
\end{figure}

On matched DAVIS, identical one-call Qwen arms intervene on edge, support, and boundary
predicates. Figure~\ref{fig:predicate-decomposition} summarizes unsafe counts; Appendix
Table~\ref{tab:edge-decision-conditions} retains every condition-level count.

Across the 560 cases, full-bank serialization yields 149 unsafe releases (26.61\%) and
typed authorization yields 5 (0.89\%). Correct releases rise from 129 to 154, while input
tokens fall from 2.032M to 0.285M.

\paragraph{Is the gain only relevance pruning?}
We replay group-level triad relevance on the same frozen full-bank candidate without
another MLLM call (Table~\ref{tab:relevance-baseline}).

\begin{table}[H]
\centering
\scriptsize
\setlength{\tabcolsep}{2.4pt}
\caption{\textbf{Relevance pruning does not match typed authorization at comparable
release} (\%). Each row uses 560 frozen Qwen calls. Relevance policies filter the same
full-bank candidate offline; \emph{Released} counts actual memory answers (useful or
unsafe). Unscorable cases fail closed; see Section~\ref{sec:limitations}.}
\label{tab:relevance-baseline}
\begin{tabular*}{\linewidth}{@{\extracolsep{\fill}}lrrrrr@{}}
\toprule
Policy & Calls & Released & Clean acc. & Aggregate unsafe & Worst-cond. unsafe \\
\midrule
Full bank & 560 & 49.64 & 67.86 & 26.61 & 92.86 \\
Top-1 relevance & 560 & 28.93 & 66.07 & 6.79 & 46.43 \\
Top-2 relevance & 560 & 38.04 & 66.07 & 15.36 & 48.21 \\
Threshold relevance & 560 & 11.07 & 25.00 & 1.07 & 10.71 \\
\textbf{Typed authorization} & \textbf{560} & \textbf{28.39} & \textbf{64.29} & \textbf{0.89} & \textbf{7.14} \\
\bottomrule
\end{tabular*}
\end{table}

Top-1 relevance and typed authorization have comparable release (28.93\% vs. 28.39\%) but
6.79\% versus 0.89\% unsafe release, with clean accuracy changing from 66.07\% to 64.29\%.
Threshold relevance is dominated: 11.07\% release, 25.00\% clean accuracy, and 1.07\% unsafe.

Of the 33 unsafe cases removed beyond top-1 relevance, support accounts for 27, edge
validity for 4, clean cases for 2, and boundary cases for 0. Cross-group duplicate
leaves one top-1 unsafe case (seven under top-2) and none under typed authorization.
$E$ and $S$ remain operationally inseparable because authorization enables the latest-usable-record trace.

On the 392 decidable cases, unsafe release falls from 90 to zero and correct release rises
from 129 to 154, primarily on support conditions (Appendix Table~\ref{tab:edge-decision-conditions}).

The 168 boundary cases retain 5 of 59 events: four target-absence events outside relative
appearance evidence (Proposition~\ref{prop:relative-boundary}) and one unidentifiable
whole-bank swap. The 52-to-1 reduction is a construction artifact because cards, but not
anchors, are permuted; permuting anchors would remove this rejection cue.

At matched false release $\leq2\%$, relative appearance raises the edge frontier from
44.64\% to 66.07\% useful-case recall; the presence frontier remains 60.71\% despite an
8.92-point localization gain (Appendix~\ref{app:typed-frontiers}). Thus $E$ is visually
decidable in the tested class, while $S$ supplies most of the gain over relevance.

\subsection{What can post-generation typed checks add?}
\label{sec:post-generation}

At fixed candidates, the 560-call margin changes unsafe release from 5 to 3 events
(0.89\% to 0.54\%) at 64.29\% clean accuracy. A second candidate reaches 73.21\% with
1,120 calls; Appendix Table~\ref{tab:cec-ablation} and Figure~\ref{fig:main-analysis}
give the descriptive, source-dependent frontier.

\subsection{What observation is missing?}

Target absence dominates the boundary: no supplied record belongs to the queried subject.
A subject-equivalent token tests the missing witness in Proposition~\ref{prop:relative-boundary}.

\begin{table}[H]
\centering
\scriptsize
\setlength{\tabcolsep}{2.2pt}
\caption{\textbf{Presence requires an external observation (\%).} Fixed candidates use
supplied subject-equivalent query/group/record tokens; DAVIS IDs are not inferred from
appearance. Zeroes are counts, not risk bounds; appearance policies are from Appendix
Table~\ref{tab:cec-ablation}.}
\label{tab:channel-frontier}
\resizebox{\linewidth}{!}{%
\begin{tabular}{llrrr}
\toprule
Panel & Observation channel & Clean accuracy & Aggregate unsafe & Worst-condition unsafe \\
\midrule
DAVIS-56 & best appearance-based & 73.21 & 0.54 & 5.36 \\
DAVIS-56 & authenticated tokens (oracle) & \textbf{85.71} & \textbf{0.00} & \textbf{0.00} \\
Support/Qwen--CoViP & authenticated tokens (oracle) & 87.30--88.93 & 0.00 & 0.00 \\
\bottomrule
\end{tabular}}
\end{table}

Authenticated tokens remove the target-absence residue (5.36\% condition-wise versus
0.54\% overall), reaching the candidate-union ceiling with 0 of 560 observed false releases.
As DAVIS track IDs, they are a sufficiency control: supplied evidence decides $E$ and $S$,
whereas $P$ requires provenance.

On free-form questions, target-absence false answers fall to 1.61\%, but attribute-absence
false answers remain 46.30--53.70\% (Appendix~\ref{app:free-form}); the unvalidated judge
indicates a predicate shift, not accuracy, and MyVLM remains a boundary audit.

\section{Limitations and Discussion}
\label{sec:limitations}

The study establishes inducibility under frozen contracts, not prevalence or realized harm.
Of 33 additional unsafe cases removed beyond relevance, 27 are support and 4 edge failures,
identifying an $E\wedge S$ pipeline rather than an edge-only mechanism.
The offline relevance replay fails closed on 19 confirmation cases with actual memory
answers---17 with a missing record image, one without a unique trace, and one without
localized query evidence. This favors the relevance baseline because unsafe answers among
those cases cannot enter its 6.79\% rate, whereas the five typed-authorization unsafe
releases are counted through the separately generated authorized path; 6.79\% is therefore
an optimistic lower bound for this replay. The replay also filters a candidate generated
from the complete bank rather than regenerating from a pruned bank; a true pre-generation
top-$k$ context may change the answer, and the direction of that difference is unknown.
The whole-bank-swap reduction relies on unpermuted anchors and is not general capability.
Localized rows condition on masks, no check infers presence, and the split is not video-disjoint.
Relative-evidence claims hold only within their observation class; typed authorization is
not visual identity authentication, and lower-capability interfaces remain boundary audits.
RecordAuth-Diag covers Qwen-, Phi-, and Gemma-family interfaces but no proprietary VLMs;
differences in scale, training data, and interface preclude a cross-model ranking. We omit MMPB because its
interface lacks the record assignments needed for matched authorization.

\vspace{-0.6em}
\section{Conclusion}

RecordAuth-Diag isolates authorization from recall; raw-bank misbinding is observed across
Qwen-, Phi-, and Gemma-family interfaces. Typed authorization lowers Gemma local
unauthorized use by 20.78 points and DAVIS unsafe use from 6.79\% to 0.89\%, mainly
through support. The full $P\wedge E\wedge S$ contract still requires authenticated presence.

\clearpage
\subsection*{AI use statement}

Generative AI tools assisted with hypothesis refinement, experimental-design feedback,
the conceptual framework, formulation and checking of mathematical claims and proofs,
method and analysis implementation, synthetic-metadata construction and cleaning,
result interpretation, translation, literature discovery and summarization, manuscript
editing, reference and table formatting, and schematic revision. They were not sources
of empirical measurements or substitutes for model inference or human annotation;
interview, survey, transcription, and qualitative thematic-analysis tasks were not
applicable. The authors verified citations against source records, regenerated reported
measurements from saved predictions, reviewed the proofs, tested AI-assisted code and
artifacts, and take responsibility for the final content.

\subsection*{Ethics statement}

This study uses public research datasets and synthetic event metadata; no new
human-subject data were collected. Public availability and the evaluated diagnostic
contracts do not confer consent, legal identity, or permission to collect personal
information. Deployment would require authenticated enrollment and updates, data
minimization, access control, deletion, demographic evaluation, and legal review. The
trusted masks and provenance tokens used in controlled analyses are supplied
observations rather than identities inferred from appearance.

\subsection*{Reproducibility statement}

Appendix~\ref{app:reproducibility} specifies checkpoint revisions, environments,
inference settings, evidence registries, and the table-to-artifact mapping. The
\artifactlink{} contains code, manifests, frozen predictions, validation grids, logs,
and table-producing analyses, including the held-out selection, counterfactual-policy ablations,
relevance-pruning replay, same-union frozen-ATRA control, path-matched HGB and MyVLM
audits, typed frontiers, and provenance decisions. Other appendix sections provide the diagnostic construction,
metrics, proofs, retained failures, and hardware details.

\bibliography{references}
\bibliographystyle{iclr2027_conference}

\appendix
\section{Typed identifiability proofs and authorization protocol}
\label{app:typed-identifiability}

\subsection{Relative-evidence non-identifiability}

Let a bank-relative policy be $a(T(v))$, where $v\in\mathbb R^J$ is a vector of
candidate evidence scores and $T(v+c\mathbf 1)=T(v)$ for every constant $c$.
Consider a present world with score vector $v$ and an absent world with vector
$v+c\mathbf 1$. The policy receives the same value of $T$ in both worlds and hence
has the same release distribution. If false release in the absent world is bounded by
$\alpha$ uniformly over this world class, release in the paired present world is also
at most $\alpha$. Rankings and pairwise margins satisfy the stated invariance. The
construction does not apply to selectors with an independently calibrated absolute
presence observation.

\subsection{Typed bottleneck bound}

For one predicate type $\ell\in\{P,E,S\}$, let $\mathbb P^+$ denote the common all-valid
observable distribution and $\mathbb P_\ell^-$ a matched distribution that flips only
predicate $\ell$ while retaining the same candidate indices. Final release implies
acceptance by gate $a_\ell$, so
\begin{align}
\Pr_{\mathbb P^+}(\mathrm{release})
&\leq \mathbb E_{\mathbb P^+}[a_\ell]\\
&\leq \mathbb E_{\mathbb P_\ell^-}[a_\ell]
   +\operatorname{TV}(\mathbb P^+,\mathbb P_\ell^-)\\
&\leq \alpha_\ell+\operatorname{TV}(\mathbb P^+,\mathbb P_\ell^-).
\end{align}
Taking the minimum over $\ell$ gives Eq.~\ref{eq:typed-bottleneck}. The middle
inequality is the standard bounded-function characterization of total variation; no
new generic TV result is claimed.

\subsection{Conditional edge guarantee}

Assume that every traced authorized candidate satisfies
$\Delta_g\geq\delta_+$ and every traced misbound candidate satisfies
$\Delta_g\leq\delta_-$, with $\delta_-<\delta_+$. For any
$\tau\in(\delta_-,\delta_+]$, thresholding the margin accepts all candidates in the
first set and none in the second. If the record trace and support witness are exact,
every accepted present-identity candidate has a valid edge and support. This guarantee
does not include target presence; an absent query may still have a positive best-group
margin.

\subsection{Authenticated-provenance sufficiency}

Assume that the query subject, group registry, and record carry authenticated tokens
$z_q,z_g,z_r$ whose equality denotes the same authorization subject. Tokens cannot be
forged or silently reassigned. Releasing a uniquely traced candidate only when
$z_q=z_g=z_r$ and the support witness succeeds implies
$P_j=E_{jk}=S_{jk}=1$ for the same cited indices, so typed false acceptance is zero.
Conversely, every correctly grounded candidate in
either generation path has matching tokens and support and is accepted. The dual-path
recall therefore equals candidate-union availability. DAVIS track identities realize
this assumption for a controlled proof of concept; they are evaluation metadata rather
than visually inferred credentials.

\subsection{Frozen confirmation protocol}

Method development consumed the 488-parent Binding$\times$Support panel, DreamBooth,
and a stable 100-track DAVIS subset. The remaining 56 DAVIS tracks were selected by set
difference before new generation; their query identities have zero overlap with the
development subset. All ten conditions contribute 56 cases, for 560 total. The raw and
guard candidates use the original crops, while edge scoring uses a derived
instance-mask view; multi-subject composites without a single localized query fail
closed. Guard/raw thresholds $0.02/0.075$, path priority, parser, and evidence assets
were hashed before Qwen inference. No held-out result changed the method.

The confirmation set contains 56 tracks from 43 source videos. Twenty-five of those
videos also contribute different object tracks to the 100-track development set, so the
split is track-disjoint but not video-disjoint. Re-clustering the clean contrast by
source video gives $[1.69,18.52]$ points; leave-one-video-out gains range from 5.56 to
9.80 points. A post-hoc sensitivity restricted to 19 tracks from 18 videos absent from
development gives a 10.53-point gain with interval $[0.00,26.32]$ and one unsafe case
for each policy.

\subsection{Predicate-level pre-generation counts}

\begin{table}[H]
\centering
\scriptsize
\setlength{\tabcolsep}{2.6pt}
\caption{\textbf{Authorization is exhausted where evidence decides, and leaks only outside
that class} (56 cases per condition; 560 calls in both arms). \emph{Corr.}: releases
answering from an authorized record; \emph{unsafe}: released answers copying an
unauthorized record. Conditions are grouped by the predicate their intervention corrupts.
The last block is outside the observation class of Section~\ref{sec:identifiability} and
provides the exact counts underlying Figure~\ref{fig:predicate-decomposition}.}
\label{tab:edge-decision-conditions}
\begin{tabular*}{\linewidth}{@{\extracolsep{\fill}}llrrrr@{}}
\toprule
& & \multicolumn{2}{c}{Full bank} & \multicolumn{2}{c}{Typed auth.} \\
\cmidrule(lr){3-4}\cmidrule(lr){5-6}
Predicate & Condition & Corr. & Unsafe & Corr. & Unsafe \\
\midrule
\multirow{3}{*}{Edge ($E$)}
 & Reciprocal edge swap & 0 & 27 & 0 & \textbf{0} \\
 & One-way replacement & 0 & 4 & 0 & \textbf{0} \\
 & Cross-group duplicate & 33 & 10 & 34 & \textbf{0} \\
\midrule
\multirow{2}{*}{Support ($S$)}
 & Stale latest record & 16 & 28 & 36 & \textbf{0} \\
 & Missing latest view & 21 & 19 & 33 & \textbf{0} \\
\midrule
\multirow{2}{*}{None}
 & Clean & 38 & 1 & 36 & \textbf{0} \\
 & Clean, reordered bank & 21 & 1 & 15 & \textbf{0} \\
\midrule
\multicolumn{2}{l}{\textbf{Decidable subtotal (392 cases)}} & \textbf{129} & \textbf{90} & \textbf{154} & \textbf{0} \\
\midrule
\multirow{3}{*}{\shortstack[l]{Outside\\obs.\ class}}
 & Target absent & 0 & 6 & 0 & 4 \\
 & Whole-bank swap & 0 & 52 & 0 & 1 \\
 & Multi-subject ambiguous & 0 & 1 & 0 & \textbf{0} \\
\midrule
\multicolumn{2}{l}{Boundary subtotal (168 cases)} & 0 & 59 & 0 & 5 \\
\midrule
\multicolumn{2}{l}{\textbf{Total (560 cases)}} & \textbf{129} & \textbf{149} & \textbf{154} & \textbf{5} \\
\bottomrule
\end{tabular*}
\end{table}

\begin{table}[H]
\centering
\small
\setlength{\tabcolsep}{3.2pt}
\caption{Frozen 56-track counterfactual-policy condition results (\%). Path rates identify where the
released candidate originated. Unsafe is nonzero only for target absence.}
\label{tab:cec-condition-full}
\resizebox{\linewidth}{!}{%
\begin{tabular}{lrrrr}
\toprule
Condition & Answer acc. & Unsafe & Guard path & Raw-rescue path \\
\midrule
Clean & 73.21 & 0.00 & 64.29 & 8.93 \\
Clean reordered & 46.43 & 0.00 & 26.79 & 19.64 \\
Cross-group duplicate & 69.64 & 0.00 & 60.71 & 8.93 \\
Latest one-way replacement & 0.00 & 0.00 & 0.00 & 0.00 \\
Latest reciprocal swap & 0.00 & 0.00 & 0.00 & 0.00 \\
Missing latest view & 57.14 & 0.00 & 55.36 & 1.79 \\
Multi-subject ambiguous & 100.00 & 0.00 & 0.00 & 0.00 \\
Stale latest record & 64.29 & 0.00 & 62.50 & 1.79 \\
Target absent & 94.64 & 5.36 & 5.36 & 0.00 \\
Whole-group swap & 100.00 & 0.00 & 0.00 & 0.00 \\
\bottomrule
\end{tabular}}
\end{table}

The paired clean gain over guard is 8.93 points with a 95\% track-cluster
bootstrap interval of $[1.79,16.07]$. Three target-absence releases among 56 cases give
5.36\% unsafe exposure (Clopper--Pearson 95\% interval $[1.12,14.87]$). Aggregate
unsafe exposure is 0.54\%. Against the aligned guard outputs, unsafe counts are 3/560
versus 5/560; the two discordant cases are both guard-only, and the track-cluster
interval for the $-0.36$-point difference is $[-0.89,0.00]$. Thus the aggregate rate
does not reveal the condition-specific presence failure, and the observed reduction is
not statistically established.

\subsection{Post-generation policy comparison}
\label{app:post-generation}

\begin{table}[H]
\centering
\scriptsize
\setlength{\tabcolsep}{2.0pt}
\caption{\textbf{Post-generation filtering trades released answers for the last few
events (\%).} All rows use the 560-case DAVIS-56 union generated once by
Qwen3-VL-8B-Instruct (one guard and one raw candidate per case). \emph{Released} is the
fraction of cases in which the policy emits a memory-derived answer rather than
\nullmem{}. Frozen ATRA is an offline no-new-call control; HGB is post-hoc; the oracle
uses labels. Aggregate unsafe counts are 5/560 for typed authorization, 3/560 with the
margin, and 7/560 for frozen ATRA; the 56 tracks span 43 source videos.}
\label{tab:cec-ablation}
\resizebox{\linewidth}{!}{%
\begin{tabular}{lrrrrr}
\toprule
Policy & Calls & Released & Clean accuracy & Aggregate unsafe & Worst-condition unsafe \\
\midrule
Raw candidate, full bank & 560 & 49.64 & 67.86 & 26.61 & 92.86 \\
\textbf{Pre-generation typed authorization} & \textbf{560} & \textbf{28.39} & \textbf{64.29} & \textbf{0.89} & \textbf{7.14} \\
\quad + counterfactual margin & 560 & 27.50 & 64.29 & 0.54 & 5.36 \\
Raw + margin, no typed authorization & 560 & 17.86 & 48.21 & 0.00 & 0.00 \\
\midrule
Dual path, schema only & 1,120 & 50.00 & 85.71 & 15.54 & 76.79 \\
Dual path, frozen ATRA filter & 1,120 & 32.32 & 71.43 & 1.25 & 10.71 \\
Dual path, post-hoc HGB & 1,120 & 32.68 & 66.07 & 0.71 & 5.36 \\
Dual margin, no localization & 1,120 & 35.71 & 78.57 & 1.25 & 10.71 \\
Dual margin + localization & 1,120 & 31.61 & 73.21 & 0.54 & 5.36 \\
\midrule
Candidate-union oracle & 1,120 & 8.57 & 85.71 & 0.00 & 0.00 \\
\bottomrule
\end{tabular}}
\end{table}

\begin{figure}[t]
  \centering
  \includegraphics[width=\linewidth]{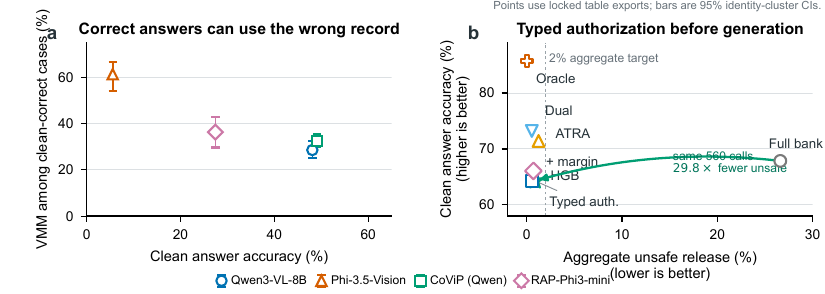}
  \caption{\textbf{Two views of the locked evidence.} (a) Among clean-correct cases,
  all four interfaces still use an unauthorized record at nontrivial rates; bars are
  95\% identity-cluster intervals. (b) On the frozen DAVIS-56 union, full-bank and
  typed-authorization points bracket the pre-generation change, while post-generation
  policies redistribute a few remaining events at reduced coverage. Main-paper
  Table~\ref{tab:relevance-baseline} supplies the relevance-matched control and attributes
  the incremental gain to $E\wedge S$, dominated by support; the oracle is an upper bound,
  and dual-path policies use twice the calls.}
  \label{fig:main-analysis}
\end{figure}

\paragraph{Matched learned selector.}
This post-hoc candidate-conditioned HGB control uses the five Attempt-4 checkpoints, the
same localized raw/guard candidate union, behavioral trace, support rule, and guard-first
priority. To match the dual-path counterfactual policy's class, guard and raw thresholds are selected separately
on DAVIS-100. Maximizing clean accuracy subject to both aggregate and maximum-condition
unsafe rates no greater than 2\% gives thresholds $0.832156/0.984147$ and a development
endpoint of 65.00/0.40/1.00\% clean/aggregate/maximum risk. Without DAVIS-56 retuning,
the control reaches 66.07/0.71/5.36\%; the counterfactual policy reaches 73.21/0.54/5.36\%. Its clean
gain is 7.14 points with track-bootstrap interval $[1.79,14.29]$. The control was
designed after DAVIS-56 inspection, although calibration uses no DAVIS-56 labels; it
therefore remains diagnostic rather than independent confirmation.

A complete post-hoc sweep over all 308 shared HGB thresholds finds no point that
simultaneously matches or exceeds the counterfactual policy in clean accuracy while weakly improving both risk
metrics. At HGB aggregate risk no greater than the counterfactual policy, clean accuracy is 58.93\%, aggregate
unsafe is 0.18\%, and maximum-condition unsafe is 1.79\%; at maximum-condition risk
matched to the counterfactual policy, HGB reaches 78.57/1.79/5.36\% clean/aggregate/maximum risk. These
test-selected endpoints diagnose a crossing frontier and are not deployable operating
points or independent evidence.

\subsection{Output-held-out MyVLM transfer audit}

The counterfactual policy, the path-specific HGB thresholds, parsers, and candidate paths were frozen before
new generation on all 1,190 MyVLM cases. The source itself is not pristine: its images
and some clean cases had appeared in earlier benchmark/authorizer work. The experiment
therefore tests newly generated candidate outputs under a frozen post-generation rule,
not independent-source confirmation. Both selectors use the frozen full-frame SigLIP2
channel; no mask annotation is introduced.

\begin{table}[t]
\centering
\small
\setlength{\tabcolsep}{3.0pt}
\caption{Output-held-out MyVLM audit (119 clean banks; 1,190 total cases). The counterfactual policy and HGB
use identical raw/guard candidates. Parentheses report the 95\% query-identity-cluster
interval for counterfactual-policy minus HGB clean accuracy.}
\label{tab:myvlm-cec-transfer}
\resizebox{\linewidth}{!}{%
\begin{tabular}{lrrrrr}
\toprule
Generator & Raw clean & Guard clean & Path-HGB clean & Counterfactual clean & Counterfactual unsafe \\
\midrule
Qwen3-VL-8B & 51.26 & 25.21 & 37.82 & 41.18 ($[-3.33,10.66]$) & 0/1,190 \\
CoViP (Qwen) & 53.78 & 36.13 & 39.50 & 47.06 ($[0.00,16.67]$) & 0/1,190 \\
\bottomrule
\end{tabular}}
\end{table}

Both raw generators fall below the registered 61\% capability floor. The counterfactual policy improves over
the guard by 15.97 points for Qwen (interval $[9.40,23.85]$) and 10.92 for CoViP
($[6.50,15.65]$), but loses 10.08 and 6.72 points relative to raw generation. Zero
unsafe events have a 0.31\% aggregate Clopper--Pearson upper endpoint; the 0/119
per-condition upper endpoint is 3.05\%. The capable-transfer gate therefore fails, as
does the predeclared matched-method gate: Qwen's HGB gain is below five points with an
interval crossing zero, and CoViP's interval touches zero. A 20-parent independent
Idefics3 preflight obtains 0\% raw clean accuracy and is not scaled.

Complete file-to-claim SHA-256 values are machine-readable in the supplementary artifact's
\texttt{SOURCE\_INVENTORY.json} and \texttt{UPLOAD\_MANIFEST.json}; they are omitted
from the typeset appendix because the identifiers carry no independent scientific
content.

\section{Free-form typed decomposition}
\label{app:free-form}

\begin{table}[H]
\centering
\scriptsize
\setlength{\tabcolsep}{2.4pt}
\caption{\textbf{Free-form questions expose an answer-support bottleneck (\%).}
False-answer endpoints use the benchmark's frozen abstention keyword contract;
answerable accuracy uses the local Phi sensitivity judge and is not human-validated.}
\label{tab:omni-typed}
\resizebox{\linewidth}{!}{%
\begin{tabular}{lrrrr}
\toprule
Run & Answerable accuracy & Target-absent false & Attribute-absent false & Support--presence gap \\
\midrule
Qwen raw & 68.70 & 8.06 & 74.07 & 66.01 \\
Qwen strict prompt & 66.09 & 1.61 & 53.70 & 52.09 \\
CoViP raw & 74.78 & 9.68 & 77.78 & 68.10 \\
CoViP strict prompt & 65.22 & 1.61 & 46.30 & 44.68 \\
\bottomrule
\end{tabular}}
\end{table}

The support--presence gap is the difference between attribute-absence and
target-absence false-answer rates. Strict prompting closes the presence side almost
completely while leaving the support side largely intact, which is why we read answer
support as the limiting predicate in this panel. The Omni-Persona protocol, prompt
variants, and judge configuration are specified in Appendix~\ref{app:omni}.

\section{Typed frontiers across panels}
\label{app:typed-frontiers}

\begin{table}[H]
\centering
\scriptsize
\setlength{\tabcolsep}{2.2pt}
\caption{\textbf{Typed frontiers reveal different observation bottlenecks (\%).}
Entries are the maximum useful-case recall at matched false release $\leq2\%$.
``Relative'' uses the counterfactual margin and ``absolute'' uses the triad score;
dashes mean that the panel does not isolate presence.}
\label{tab:typed-frontier}
\begin{tabular*}{\linewidth}{@{\extracolsep{\fill}}llrrrr@{}}
\toprule
Panel & Candidate path & Presence, rel. & Presence, abs. & Edge, rel. & Edge, abs. \\
\midrule
Binding$\times$Support/Qwen & raw & -- & -- & 75.20 & 68.24 \\
Binding$\times$Support/CoViP & raw & -- & -- & 76.23 & 69.06 \\
DreamBooth/Qwen & raw & 70.00 & 83.33 & 86.67 & 83.33 \\
DAVIS-100 localized & raw & 49.00 & 36.00 & 64.00 & 45.00 \\
DAVIS-56 unlocalized & raw & 51.79 & 50.00 & 60.71 & 28.57 \\
DAVIS-56 localized & raw & 60.71 & 57.14 & 66.07 & 44.64 \\
DAVIS-56 localized & guard & 62.50 & 62.50 & 64.29 & 64.29 \\
RAP-Phi3 & raw & 14.95 & 17.39 & 25.54 & 20.65 \\
\bottomrule
\end{tabular*}
\end{table}

The relative-evidence column exceeds the absolute-score column for the edge predicate on
every panel except the guard-path DAVIS row, where the two coincide because the
pre-generation decision has already removed the candidates that the margin would reject.
The presence columns show no comparable separation, which is the empirical counterpart of
Proposition~\ref{prop:relative-boundary}.

\section{Identifiability of whole-bank ownership}
\label{app:identifiability}

Let the unanchored input be an unordered collection of opaque labels and mutable groups, $X=\{(s_j,M_j)\}_{j=1}^{J}$, where the labels have no semantics outside the bank. Let $\pi$ be a non-identity permutation and consider a world in which the intended owners of the groups are permuted by $\pi$. Because no observation independently ties $s_j$ to a visual subject, relabeling the opaque names by $\pi^{-1}$ maps the second world back to the same $X$. The two ownership functions disagree while the observable input is isomorphic. Thus the intended ownership is not identifiable from $X$ alone. An enrollment anchor adds an observation $a_j$ whose association with $s_j$ is fixed across the intervention, breaking this symmetry. This argument establishes necessity of side information for the stated global swap; it does not establish that one anchor is sufficient under spoofing or severe appearance change.

\section{Condition taxonomy and invariants}

\begin{table}[t]
\centering
\small
\caption{Evidence sources are kept separate. ``Cases'' are derived evaluation cases; repeated conditions or frames are not independent identities.}
\label{tab:scope}
\setlength{\tabcolsep}{3.5pt}
\begin{tabularx}{\linewidth}{@{}lrrrr>{\raggedright\arraybackslash}X@{}}
\toprule
Source & IDs/tracks & Parents & Cases & E2E cases & Role \\
\midrule
RecordAuth-Diag (LSD+Yo'LLaVA) & 87 & 369 & 3,690 & 3,690 & five-fold dev/test \\
MyVLM & 29 & 119 & 1,190 & -- & observed transfer \\
DreamBooth & 30 & 30 & 300 & 300 & prospective positive \\
DAVIS-2017 & 156 & 156 & 1,560 & 1,000 & prospective same-instance \\
POPE/COCO & 500 images & 500 & 500 & 500 & external negative \\
MMVP & 150 pairs & 150 & 300 & 300 & external negative \\
\bottomrule
\end{tabularx}
\end{table}

Every causal parent has the same question schema and a registered answer-bearing event. Clean reordering changes only card order. Reciprocal and one-way interventions move the latest visual event between identity groups while retaining the underlying images and records. Whole-group swap moves all mutable cards but not the anchors. Cross-group duplication copies an edge into a second plausible group. Missing and stale conditions preserve the record while changing image availability or validity metadata. Target absence removes a licensed target from the candidate bank, and multi-subject ambiguity makes more than one visual subject plausible. Builder audits verify case identifiers, expected answers, group/card bounds, anchor immutability, and exact/pHash duplicate constraints before any model inference.

The DAVIS source uses the sorted union of the official 2017 train and validation
videos. A track is eligible only with at least eight annotated views satisfying fixed
mask-area, box-size, border-truncation, temporal-gap, and duplicate constraints.
Before any DAVIS model output, the end-to-end subset was defined as the first 100
eligible tracks under stable hash \texttt{davis2017-e2e-v1}; all ten cases of each
selected track are retained. No gate score, prediction, or generator output enters
eligibility or ordering.

\section{Formal endpoints and output attribution}
\label{app:metrics}

Let $N$ be a reported case set, $P=\{i:y_i\neq\nullmem\}$ its positive cases,
$Z=\{i:y_i=\nullmem\}$ its null cases, $S$ the eight non-clean stress conditions,
$c_i$ exact answer correctness, $a_i$ generator abstention, $g_i$ non-null gate
authorization, and $u_i$ the relevant unsafe indicator. We compute
\begin{align*}
\mathrm{Acc}(N)&=|N|^{-1}\sum_{i\in N}c_i,&
\mathrm{Coverage}(N)&=|N|^{-1}\sum_{i\in N}g_i,\\
\mathrm{PosRecall}&=|P|^{-1}\sum_{i\in P}c_i,&
\mathrm{NullSpec}&=|Z|^{-1}\sum_{i\in Z}a_i,\\
\mathrm{Unsafe}(N)&=|N|^{-1}\sum_{i\in N}u_i,&
\mathrm{StressUnsafe}&=|S|^{-1}\sum_{i\in S}u_i.
\end{align*}
Clean recall uses the two clean conditions. At the gate, $u_i=1$ when any exposed
event lies outside $E^\star_i$; at the generator, it is one when the parsed answer
matches at least one supplied event and none of the matching events is authorized.
Thus ``unsafe exposure'' and ``unsafe use'' are distinct. For generator-frontier
tables, response coverage is reported explicitly as $1-|N|^{-1}\sum_i a_i$ rather
than conflated with gate coverage. Conditional VMM retains the eight stress cases only
for parents whose clean case is correct.

The parser takes the last \verb|\boxed{...}| span, or otherwise the final nonempty
line. Case and punctuation are normalized; a location matches either its complete
synthetic label or the label with its four-hex disambiguating suffix omitted.
Abstention recognizes \texttt{CANNOT DETERMINE} and a fixed list of equivalent
insufficiency phrases. Counterfactual-attribution headline metrics are stricter: the
complete relabeled suffix must match, so parser aliases cannot create a nonce copy.

\begin{table*}[t]
\centering
\scriptsize
\setlength{\tabcolsep}{2.2pt}
\caption{\textbf{Typed authorization reduces unauthorized use on Gemma-3-4B-IT} (\%). All arms use
the same 3,690 RecordAuth-Diag cases, checkpoint, and frozen format-tolerant parser.
Raw supplies the full bank; group/\nullmem{} supplies one routed group and bypasses
generation at \nullmem{}; typed edge additionally verifies individual cards. Brackets
are 95\% query-identity-cluster intervals. Group routing alone does not reduce local
unauthorized use. Whole-bank ownership remains outside the identifiable observation class.}
\label{tab:gemma-arms}
\resizebox{\textwidth}{!}{%
\begin{tabular}{lrrrr}
\toprule
Arm & Clean recall [CI] & Local unauthorized [CI] & Whole-bank swap [CI] & Unauthorized $\mid$ clean-correct [CI] \\
\midrule
Raw full bank & 25.75 [22.62,29.23] & 63.69 [59.74,67.84] & 100.00 [100.00,100.00] & 59.29 [54.50,63.83] \\
Group/\nullmem{} & 28.73 [25.52,31.68] & 73.35 [69.35,76.73] & 96.75 [94.56,98.56] & 62.39 [58.65,65.68] \\
Group + typed edge & 25.75 [22.48,28.76] & 42.91 [38.59,47.72] & 82.66 [77.90,86.87] & 15.73 [11.29,20.28] \\
\bottomrule
\end{tabular}}
\end{table*}

\begin{table*}[t]
\centering
\scriptsize
\setlength{\tabcolsep}{2.3pt}
\caption{Raw unauthorized-memory use by intervention (\%, 95\% query-identity-cluster bootstrap CI). All rows use the frozen native interfaces; RAP excludes one integration-smoke parent. Local corruptions remain substantial and are reported separately from the whole-group swap ($^\ddagger$), which is structurally unidentifiable without invariant side information.}
\label{tab:all-condition-raw}
\resizebox{\textwidth}{!}{%
\begin{tabular}{lrrrr}
\toprule
Intervention & Qwen3-VL & Phi-3.5 & CoViP & RAP-Phi3 \\
\midrule
Clean & 2.17 [0.75,3.87] & 27.37 [21.71,33.33] & 2.98 [1.38,4.80] & 2.17 [0.82,3.93] \\
Clean reordered & 1.08 [0.25,2.24] & 23.85 [19.59,28.27] & 1.63 [0.53,2.90] & 2.17 [0.82,3.88] \\
Target absent & 2.17 [0.58,4.17] & 41.19 [35.03,47.48] & 2.98 [1.04,5.59] & 93.21 [90.68,95.54] \\
Reciprocal swap & 40.92 [35.20,46.96] & 33.06 [27.42,38.98] & 46.61 [42.01,51.63] & 26.63 [21.79,32.28] \\
One-way replacement & 12.74 [9.58,16.31] & 34.42 [28.75,40.65] & 16.26 [13.03,19.58] & 9.78 [6.47,13.95] \\
Cross-group duplicate & 13.82 [10.26,17.61] & 25.75 [20.65,31.03] & 15.18 [11.75,18.76] & 22.55 [18.10,27.34] \\
Missing latest view & 27.64 [22.71,32.76] & 49.32 [43.97,54.70] & 31.98 [26.78,37.50] & 9.24 [6.08,13.23] \\
Stale latest record & 45.80 [39.30,52.27] & 48.51 [42.71,54.64] & 48.51 [42.53,54.59] & 35.60 [30.23,41.71] \\
Multi-subject ambiguous & 1.36 [0.29,2.55] & 15.72 [12.13,19.69] & 2.44 [0.98,4.11] & 1.09 [0.25,2.30] \\
\midrule
Whole-group swap$^\ddagger$ & 90.79 [87.20,93.98] & 41.73 [36.13,47.76] & 94.31 [91.76,96.63] & 97.83 [95.44,99.67] \\
\bottomrule
\end{tabular}}
\end{table*}

\begin{table*}[t]
\centering
\scriptsize
\setlength{\tabcolsep}{2.2pt}
\caption{Generator-level causal attribution on the population fixed by each model's frozen raw-bank unsafe outputs (\%). Brackets are 95\% query-identity-cluster intervals; an all-zero resampling interval is not a population upper bound. `Same' requires suffix-preserving equality with the raw unsafe answer. `Copy' matches any displayed intervened label; shuffle preserves the label multiset, so its event column is the specific tracking test. Nonce labels occur only after intervention. `Event' requires the new label to remain attached to the event selected in the raw run.}
\label{tab:memory-attribution-full}
\resizebox{\textwidth}{!}{%
\begin{tabular}{lrrrrrrr}
\toprule
Model & Raw-unsafe $n$ & No-memory same & Shuffle any-label & Shuffle event & Nonce copy & Nonce event & Nonce unauth. copy \\ 
\midrule
Qwen3-VL-8B & 880 & 0.00 [0.00,0.00] & 87.50 [84.86,89.92] & 66.14 [62.14,70.36] & 86.59 [82.02,90.84] & 57.95 [53.11,63.04] & 70.68 [66.36,74.90] \\
Phi-3.5-Vision & 1258 & 0.00 [0.00,0.00] & 14.23 [11.55,17.05] & 3.74 [2.31,5.44] & 60.33 [55.35,65.38] & 15.42 [12.68,18.33] & 53.90 [49.17,58.63] \\
CoViP (Qwen) & 970 & 0.00 [0.00,0.00] & 97.22 [95.78,98.37] & 71.86 [68.29,75.63] & 92.78 [89.15,95.98] & 65.26 [61.04,69.88] & 78.35 [74.29,82.29] \\
\bottomrule
\end{tabular}%
}
\end{table*}
\begin{table*}[t]
\centering
\scriptsize
\setlength{\tabcolsep}{2.5pt}
\caption{Parser and output-format sensitivity on each frozen raw-unsafe population (\%). Exact matching retains the complete synthetic suffix; compatible matching mirrors the diagnostic parser. Format failure requires deviation from one single-line boxed answer. `Unrecognized' is a non-abstaining answer matching neither an original-bank nor displayed-prompt location. Brackets are 95\% query-identity-cluster intervals.}
\label{tab:attribution-parser}
\resizebox{\textwidth}{!}{%
\begin{tabular}{llrrrr}
\toprule
Model & Intervention/estimand & Exact & Parser-compatible & Format failure & Unrecognized non-\nullmem{} \\ 
\midrule
\multirow{3}{*}{Qwen3-VL-8B} & No memory: raw persistence & 0.00 [0.00,0.00] & 0.00 [0.00,0.00] & 19.77 [13.54,26.27] & 1.48 [0.12,3.40] \\
 & Shuffle: prompt copy & 87.50 [84.86,89.92] & 96.25 [94.89,97.48] & 7.84 [5.04,10.71] & 0.00 [0.00,0.00] \\
 & Nonce: prompt copy & 86.59 [82.02,90.84] & 86.59 [81.97,90.74] & 6.36 [4.20,8.69] & 0.91 [0.36,1.55] \\
\midrule
\multirow{3}{*}{Phi-3.5-Vision} & No memory: raw persistence & 0.00 [0.00,0.00] & 0.00 [0.00,0.00] & 100.00 [100.00,100.00] & 16.45 [9.87,22.97] \\
 & Shuffle: prompt copy & 14.23 [11.55,17.05] & 72.34 [68.13,76.32] & 100.00 [100.00,100.00] & 6.76 [4.89,8.87] \\
 & Nonce: prompt copy & 60.33 [55.35,65.38] & 60.33 [55.23,65.49] & 100.00 [100.00,100.00] & 0.48 [0.16,0.83] \\
\midrule
\multirow{3}{*}{CoViP (Qwen)} & No memory: raw persistence & 0.00 [0.00,0.00] & 0.00 [0.00,0.00] & 45.15 [35.35,54.90] & 0.93 [0.00,2.83] \\
 & Shuffle: prompt copy & 97.22 [95.78,98.37] & 97.63 [96.25,98.74] & 83.30 [79.05,87.35] & 0.00 [0.00,0.00] \\
 & Nonce: prompt copy & 92.78 [89.15,95.98] & 92.78 [89.22,95.99] & 92.68 [90.10,95.04] & 0.21 [0.00,0.53] \\
\bottomrule
\end{tabular}%
}
\end{table*}
\begin{table*}[t]
\centering
\scriptsize
\setlength{\tabcolsep}{2.2pt}
\caption{Frozen-generator frontier endpoints on the same 3,690 RecordAuth-Diag cases (\%). These are diagnostic endpoints, not deployable methods. Oracle-authorized supplies ground-truth-applicable cards but leaves generation unchanged. Brackets are 95\% query-identity-cluster intervals.}
\label{tab:frontier-endpoints}
\resizebox{\textwidth}{!}{%
\begin{tabular}{llrrrrr}
\toprule
Model & Context & Accuracy & Coverage & Pos. recall & \nullmem{} spec. & Stress unsafe \\ 
\midrule
\multirow{3}{*}{Qwen3-VL-8B} & Raw bank & 40.16 [38.06,42.31] & 76.50 [73.21,79.63] & 37.05 [33.86,40.25] & 47.43 [44.98,49.96] & 29.40 [27.20,31.79] \\
 & No memory (actual) & 29.57 [29.15,29.89] & 1.57 [0.30,3.18] & 0.00 [0.00,0.00] & 98.55 [97.19,99.65] & 0.00 [0.00,0.00] \\
 & Oracle-authorized (actual) & 64.44 [60.05,68.89] & 51.44 [46.54,56.28] & 49.83 [43.46,56.26] & 98.55 [97.19,99.64] & 0.00 [0.00,0.00] \\
\midrule
\multirow{3}{*}{Phi-3.5-Vision} & Raw bank & 19.73 [17.94,21.44] & 50.38 [45.65,55.20] & 5.30 [3.99,6.71] & 53.39 [48.24,58.37] & 36.21 [31.72,40.90] \\
 & No memory (actual) & 26.18 [24.68,27.63] & 15.39 [9.70,21.58] & 0.00 [0.00,0.00] & 87.26 [82.16,92.06] & 0.00 [0.00,0.00] \\
 & Oracle-authorized (actual) & 67.64 [64.18,71.27] & 69.70 [67.49,71.80] & 59.23 [54.53,64.48] & 87.26 [82.20,92.04] & 0.00 [0.00,0.00] \\
\midrule
\multirow{3}{*}{CoViP (Qwen)} & Raw bank & 40.41 [38.53,42.31] & 79.32 [76.77,81.68] & 37.90 [35.21,40.60] & 46.25 [44.11,48.54] & 32.28 [30.12,34.56] \\
 & No memory (actual) & 29.76 [29.42,30.00] & 0.81 [0.00,2.12] & 0.00 [0.00,0.00] & 99.19 [98.04,100.00] & 0.00 [0.00,0.00] \\
 & Oracle-authorized (actual) & 70.84 [66.80,74.83] & 57.89 [54.09,61.49] & 58.69 [53.05,64.62] & 99.19 [98.08,100.00] & 0.00 [0.00,0.00] \\
\bottomrule
\end{tabular}%
}
\end{table*}

\begin{figure}[t]
  \centering
  \includegraphics[width=0.78\linewidth]{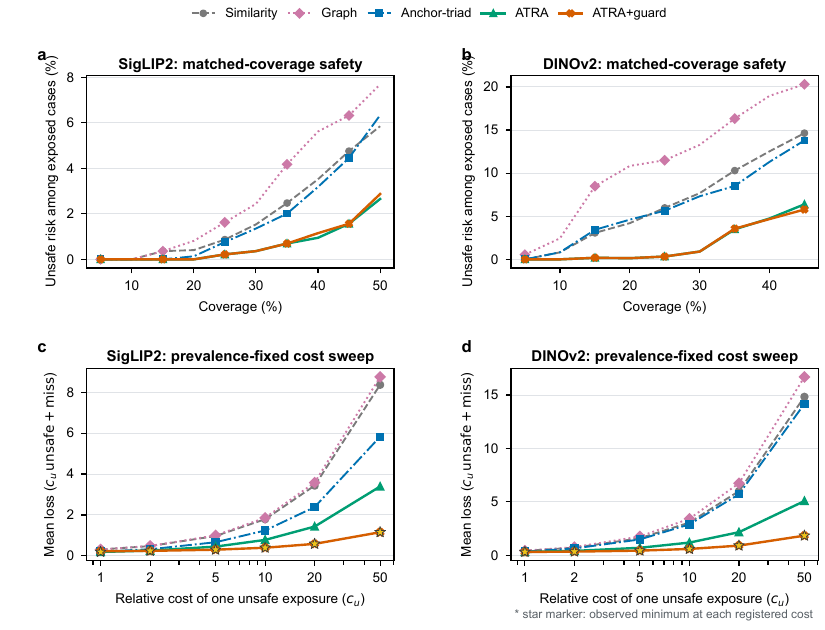}
  \caption{\textbf{Safety is selective, not free.} At matched coverage (top), learned
  relational authorization lowers unsafe risk most clearly on DAVIS; under
  cost-sensitive loss (bottom), the preferred operating point changes with the cost
  of an unsafe exposure. The figure reports a frontier rather than a universal winner.}
  \label{fig:frontier}
\end{figure}

Nonce exact and parser-compatible copy rates coincide up to bootstrap Monte Carlo
variation because the eight-character counterfactual suffix is retained. Phi and
CoViP frequently violate the requested one-line boxed format, but the headline still
requires their deterministically parsed answer to equal the complete displayed nonce;
format noncompliance therefore cannot manufacture an out-of-bank label. Human
semantic validation remains pending as stated in Appendix~\ref{app:additional-limits}.
The actual no-memory endpoint eliminates positive recall, whereas oracle-authorized
context reaches zero measured stress-unsafe use and substantially higher accuracy.
Both are diagnostic bounds rather than deployable competitors.

\section{Generator prompt and serialization audit}
\label{app:generator-prompts}

All generator-side prompting arms use the same ordered serialization: an introductory
sentence stating that subject IDs are opaque; for each group, its label followed by
its cards; then the current query image; then the question and one-line output
constraint. Each card is serialized as ``Group $g$, memory card $k$. Registered
subject: [opaque ID]. Timestamp: [value]. Record status: [active/stale]. Personal
record: [opaque ID] was seen at [location].'' A missing image is replaced by the
literal marker ``[This card has no attached visual evidence.]''

The natural question is: ``Where was the subject shown in this image most recently,
according to my visual memory?'' The verification arm appends: ``Use a record only if
its visual evidence verifies that it belongs to the shown subject. If no single memory
group is verified, answer exactly NULL.'' Both arms then append: ``Treat a missing
image or a record marked stale as unusable. Do not explain. Your entire response must
be exactly one line: \texttt{\textbackslash boxed\{<final location or NULL>\}}.''
In visible-anchor arms, exactly one anchor image precedes the cards in each group and
is introduced as ``Trusted enrollment anchor (identity reference only; not a personal
event).'' Generic interfaces interleave each text marker and image; Phi receives the
same order through numbered image placeholders. The executable formatter is included
in the artifact.

\begin{table}[t]
\centering
\small
\caption{Can the generator verify the binding itself? (\%). Visible arms add one trusted non-event anchor per group but filter no cards. Local stress pools three identifiable mismatches; conditional \vmm{} uses each arm's clean-correct parents (count in parentheses); \nullmem{} specificity pools target absence and ambiguity.}
\label{tab:same-information}
\setlength{\tabcolsep}{2.8pt}
\begin{tabular}{@{}llrrrrr@{}}
\toprule
Model & Generator view & Clean recall & Local unsafe & Cond. \vmm{} & Whole swap & \nullmem{} spec. \\
\midrule
\multirow{3}{*}{Qwen3-VL-8B} & raw, anchors hidden & 48.10 & 22.49 & 28.57 (203) & 90.79 & 66.53 \\
 & anchors, natural & 53.39 & 23.85 & 30.29 (219) & 83.47 & 65.85 \\
 & anchors, verify & 50.14 & 23.22 & 32.35 (203) & 80.49 & 77.91 \\
\midrule
\multirow{3}{*}{CoViP} & raw, anchors hidden & 49.05 & 26.02 & 32.55 (212) & 94.31 & 66.53 \\
 & anchors, natural & 54.47 & 27.64 & 35.53 (228) & 89.97 & 66.26 \\
 & anchors, verify & 54.07 & 27.91 & 35.43 (223) & 84.28 & 65.31 \\
\bottomrule
\end{tabular}
\end{table}

\section{Authorizer features and architecture}
\label{app:authorizer}

The 24 input features, in saved checkpoint order, are: query--memory cosine, anchor--memory cosine, triad minimum, query--anchor cosine, the product of query-- and anchor--memory cosine, their disagreement, group query mean and maximum, group anchor mean and maximum, group triad mean and maximum, within-group mean/minimum/maximum, maximum cross-group duplicate score, group query and triad margins, edge query margin, missing-image indicator, stale-record indicator, normalized temporal rank, group cardinality, and bank group cardinality. Binary and cardinality features are not standardized; continuous features use training-fold moments only.

The edge encoder is a two-layer 64-dimensional GELU MLP. Masked edge mean and maximum feed a two-layer group encoder. Masked group mean and maximum form global context. The group and edge heads condition on local and global states, while a separate null head scores the global state. There are 55,427 trainable parameters. On one isolated CPU thread after embeddings are available, feature construction, forward inference, and decision logic take 1.089, 0.188, and 0.053 milliseconds per case, respectively. These numbers exclude image encoding and MLLM generation and are not advertised as total system latency.

\begin{table}[t]
\centering
\small
\caption{Relational-feature ablations on all held-out RecordAuth-Diag identities (\%). No best-seed selection is used.}
\label{tab:ablation}
\begin{tabular}{lrrrrr}
\toprule
Arm & Accuracy & Unsafe & Coverage & Pos. recall & \nullmem{} spec. \\
\midrule
Anchor-triad & 74.01 & 11.46 & 72.33 & 79.09 & 62.15 \\
Query-only ATRA & 74.07 & 11.25 & 66.12 & 76.11 & 69.29 \\
ATRA w/o anchor & 70.16 & 12.47 & 65.15 & 70.81 & 68.65 \\
ATRA w/o peer & 83.06 & 9.19 & 69.05 & 84.32 & 80.13 \\
\method{} & 80.46 & 6.53 & 65.75 & 79.40 & 82.93 \\
\guard{} & 77.24 & 1.90 & 53.25 & 69.14 & 96.12 \\
\bottomrule
\end{tabular}
\end{table}

\section{Training and selection protocol}

Identity folds are fixed before feature extraction. In each fold, \method{} trains on four conditions only: clean, target absent, latest reciprocal swap, and multi-subject ambiguity. Five seeds (1, 7, 21, 42, 89) use AdamW with learning rate $10^{-3}$, weight decay $10^{-4}$, batch size 64, early stopping patience 20, and a maximum of 200 epochs. Calibration searches group-probability, group-margin, and edge thresholds on seen validation conditions. The selected point maximizes macro-condition accuracy minus twice unsafe exposure subject to retaining at least 95\% of the best clean validation accuracy. No test or prospective-source outcome changes this rule.

The \method{} threshold grid uses seven equally spaced empirical quantiles of validation group probability, group margin, and observed-edge probability, augmented by fixed boundary values $0$, $(-1,0)$, and $(0,0.5,1.000001)$, respectively. Flat-LogReg uses five quantiles and the same boundary conventions, with an additional \nullmem{}-probability grid augmented by $(0,0.5,1.000001)$. Each saved validation-grid CSV records every evaluated point and the deterministic tie-breaking fields.

\begin{table}[t]
\centering
\small
\caption{ATRA training-seed stability (\%). Each seed pools all five held-out identity folds; SD is across five complete fold replicates. The final column is the preregistered five-seed decision ensemble, not the best seed.}
\label{tab:seed-stability}
\begin{tabular}{llrrr}
\toprule
Encoder & Metric & Seed mean $\pm$ SD & Range & Ensemble \\
\midrule
SigLIP2-base & Accuracy & 78.09 $\pm$ 1.14 & 76.83--79.59 & 80.46 \\
SigLIP2-base & Unsafe & 8.09 $\pm$ 0.88 & 6.72--9.13 & 6.53 \\
SigLIP2-base & Positive recall & 77.89 $\pm$ 2.37 & 75.34--80.87 & 79.40 \\
SigLIP2-base & \nullmem{} specificity & 78.55 $\pm$ 1.95 & 76.60--81.12 & 82.93 \\
DINOv2-base & Accuracy & 67.20 $\pm$ 0.66 & 66.61--68.05 & 68.29 \\
DINOv2-base & Unsafe & 10.56 $\pm$ 1.35 & 8.62--11.76 & 9.73 \\
DINOv2-base & Positive recall & 64.20 $\pm$ 0.97 & 63.14--65.70 & 65.31 \\
DINOv2-base & \nullmem{} specificity & 74.20 $\pm$ 2.67 & 70.01--77.15 & 75.25 \\
\bottomrule
\end{tabular}
\end{table}

Table~\ref{tab:seed-stability} aggregates each seed across all five held-out folds,
rather than treating folds or cases as independent training repetitions. The final
ensemble improves over the seed mean on both encoders, but no best-seed result is used
in any headline table.

\section{Generalization to held-out intervention families}

\begin{table}[t]
\centering
\scriptsize
\setlength{\tabcolsep}{3pt}
\caption{Condition-level generalization on held-out fold identities (\%). Optimization-seen mechanisms are clean, target absence, reciprocal swap, and multi-subject ambiguity; the other six are absent from ATRA training and threshold selection but were known during feature design. This is a fixed semantic partition, not an outcome-selected split.}
\label{tab:condition-generalization}
\begin{tabular}{lllrrrr}
\toprule
Encoder & Mechanisms & Method & $n$ & Accuracy & Unsafe & Coverage \\
\midrule
SigLIP2-base & seen & Similarity & 1476 & 62.94 & 12.20 & 56.98 \\
SigLIP2-base & seen & Anchor-triad & 1476 & 65.38 & 14.02 & 64.43 \\
SigLIP2-base & seen & \method{} & 1476 & 90.04 & 3.32 & 47.83 \\
SigLIP2-base & seen & \guard{} & 1476 & 84.08 & 2.30 & 39.30 \\
SigLIP2-base & unseen & Similarity & 2214 & 64.72 & 19.24 & 84.69 \\
SigLIP2-base & unseen & Anchor-triad & 2214 & 79.77 & 9.76 & 77.60 \\
SigLIP2-base & unseen & \method{} & 2214 & 74.07 & 8.67 & 77.69 \\
SigLIP2-base & unseen & \guard{} & 2214 & 72.67 & 1.63 & 62.56 \\
DINOv2-base & seen & Similarity & 1476 & 41.26 & 30.49 & 77.10 \\
DINOv2-base & seen & Anchor-triad & 1476 & 41.06 & 32.45 & 85.37 \\
DINOv2-base & seen & \method{} & 1476 & 79.81 & 5.42 & 43.43 \\
DINOv2-base & seen & \guard{} & 1476 & 78.12 & 3.52 & 37.06 \\
DINOv2-base & unseen & Similarity & 2214 & 57.32 & 28.73 & 85.05 \\
DINOv2-base & unseen & Anchor-triad & 2214 & 66.49 & 25.38 & 87.08 \\
DINOv2-base & unseen & \method{} & 2214 & 60.61 & 12.60 & 72.09 \\
DINOv2-base & unseen & \guard{} & 2214 & 62.65 & 2.89 & 55.92 \\
\bottomrule
\end{tabular}
\end{table}

All rows in Table~\ref{tab:condition-generalization} use held-out fold identities;
``seen'' and ``held out'' refer only to optimization and threshold selection; all
families were known during feature design. On the six held-out
mechanisms with SigLIP2, \method{} changes anchor-triad accuracy/unsafe exposure
from 79.77/9.76\% to 74.07/8.67\%, while the guard reaches 72.67/1.63\%.
With DINOv2 the corresponding values are 66.49/25.38\%, 60.61/12.60\%, and
62.65/2.89\%. Thus relational authorization transfers most clearly in safety;
it does not establish accuracy dominance over the deterministic control.

\section{Matched baselines}

\begin{table}[t]
\centering
\small
\caption{Capacity-matched fixed-aggregation control on all 3,690 held-out cases (\%). Flat-MLP matches \method{}'s 24 features, folds, seeds, training, and validation-only selection, but fixes raw-feature mean/max pooling. Counts are trainable parameters.}
\label{tab:capacity-matched}
\setlength{\tabcolsep}{2.8pt}
\begin{tabular}{@{}llrrrrrr@{}}
\toprule
Encoder & Method & Params & Acc. $\uparrow$ & Unsafe $\downarrow$ & Cov. & Pos. recall $\uparrow$ & \nullmem{} spec. $\uparrow$ \\
\midrule
\multirow{3}{*}{SigLIP2} & Anchor-triad & 0 & 74.01 & 11.46 & 72.33 & 79.09 & 62.15 \\
 & \method{} & 55,427 & 80.46 & 6.53 & 65.75 & 79.40 & 82.93 \\
 & Flat-MLP & 55,773 & 81.25 & 6.91 & 67.40 & 81.15 & 81.48 \\
\midrule
\multirow{3}{*}{DINOv2} & Anchor-triad & 0 & 56.31 & 28.21 & 86.40 & 69.53 & 25.47 \\
 & \method{} & 55,427 & 68.29 & 9.73 & 60.62 & 65.31 & 75.25 \\
 & Flat-MLP & 55,773 & 65.34 & 11.63 & 59.11 & 60.98 & 75.52 \\
\bottomrule
\end{tabular}
\end{table}

\begin{table}[t]
\centering
\small
\caption{External method isolation at 30\% matched population coverage (\%). Entries are Anchor-triad/\method{}. A common frozen label-free ranking only removes native exposures; native coverage records the unrecoverable utility boundary.}
\label{tab:external-matched-coverage}
\setlength{\tabcolsep}{3.2pt}
\begin{tabular}{@{}llrrrr@{}}
\toprule
Source & Encoder & Native cov. & Unsafe@30 & Error@30 & Pos. recall@30 \\
\midrule
DreamBooth & SigLIP2 & 80.67/71.67 & 0.00/0.00 & 0.00/0.00 & 42.86/42.86 \\
DreamBooth & DINOv2 & 91.33/72.33 & 2.22/2.22 & 4.44/3.33 & 40.95/41.43 \\
\midrule
DAVIS-2017 & SigLIP2 & 77.18/71.15 & 2.56/0.43 & 2.99/0.64 & 41.58/42.58 \\
DAVIS-2017 & DINOv2 & 95.64/69.42 & 5.13/0.85 & 9.19/2.35 & 38.92/41.85 \\
\bottomrule
\end{tabular}
\end{table}

\label{app:baselines}

All deterministic thresholds are selected from the same fold validation identities. Similarity ranks groups by query--card evidence. Medoid uses the card with greatest within-group support. Graph joins cards above pairwise consistency thresholds and scores coherent components. Anchor-only scores anchor--card consistency. Anchor-triad jointly thresholds query--anchor, query--card, and anchor--card evidence and returns \nullmem{} when no candidate satisfies the rule. The appendix tables in the final evidence export include every deterministic method, validation grid, and both encoders; main-text omissions are for readability rather than selective reporting.

\begin{table}[t]
\centering
\scriptsize
\setlength{\tabcolsep}{2.8pt}
\caption{Complete deterministic, flat learned, set learned, and guarded authorizer comparison on all held-out RecordAuth-Diag identities (\%).}
\label{tab:gate-full}
\begin{tabular}{llrrrrr}
\toprule
Encoder & Method & Accuracy & Unsafe & Coverage & Pos. recall & \nullmem{} spec. \\
\midrule
\multirow{8}{*}{SigLIP2} & Similarity & 64.01 & 16.42 & 73.60 & 72.71 & 43.72 \\
 & Medoid & 61.17 & 19.81 & 74.25 & 69.45 & 41.82 \\
 & Graph & 63.82 & 17.21 & 73.33 & 72.51 & 43.54 \\
 & Anchor-only & 56.34 & 31.84 & 80.14 & 62.14 & 42.82 \\
 & Anchor-triad & 74.01 & 11.46 & 72.33 & 79.09 & 62.15 \\
 & Flat-LogReg & 72.28 & 17.53 & 83.12 & 82.66 & 48.06 \\
 & \method{} & 80.46 & 6.53 & 65.75 & 79.40 & 82.93 \\
 & \guard{} & 77.24 & 1.90 & 53.25 & 69.14 & 96.12 \\
\midrule
\multirow{8}{*}{DINOv2} & Similarity & 50.89 & 29.43 & 81.87 & 63.07 & 22.49 \\
 & Medoid & 47.37 & 35.20 & 85.66 & 60.05 & 17.80 \\
 & Graph & 49.00 & 33.20 & 85.04 & 62.37 & 17.80 \\
 & Anchor-only & 46.53 & 39.46 & 89.95 & 59.43 & 16.44 \\
 & Anchor-triad & 56.31 & 28.21 & 86.40 & 69.53 & 25.47 \\
 & Flat-LogReg & 53.93 & 30.73 & 88.05 & 67.98 & 21.14 \\
 & \method{} & 68.29 & 9.73 & 60.62 & 65.31 & 75.25 \\
 & \guard{} & 68.83 & 3.14 & 48.37 & 58.65 & 92.59 \\
\bottomrule
\end{tabular}
\end{table}
\begin{table}[t]
\centering
\small
\caption{Full \method{} versus deliberately simpler same-feature Flat-LogReg, in percentage points. Inputs, folds, conditions, and selection objective are matched, but parameter capacity is not. Intervals resample the 87 query identities. Positive values favor \method{} for accuracy/recall; negative values favor it for unsafe exposure.}
\label{tab:flat-logreg}
\begin{tabular}{llr}
\toprule
Encoder & Metric & \method{} $-$ Flat-LogReg [95\% CI] \\
\midrule
SigLIP2-base & Accuracy & +8.18 [+5.32, +10.94] \\
SigLIP2-base & Unsafe & -11.00 [-13.34, -8.68] \\
SigLIP2-base & Positive recall & -3.25 [-7.39, +0.81] \\
DINOv2-base & Accuracy & +14.36 [+11.20, +17.53] \\
DINOv2-base & Unsafe & -21.00 [-23.98, -18.11] \\
DINOv2-base & Positive recall & -2.67 [-6.39, +1.07] \\
\bottomrule
\end{tabular}
\end{table}

\section{Additional completed gate evidence}
\label{app:additional-gates}

On prospective DreamBooth, SigLIP2 anchor-triad/\method{}/guard accuracy is 84.00/93.67/95.00\% and unsafe exposure is 7.67/4.00/0.67\%. DINOv2 reaches 68.33/82.00/88.33\% accuracy and 21.67/10.33/3.00\% unsafe exposure. On all 156 DAVIS tracks, SigLIP2 reaches 69.74/83.59/84.62\% accuracy and 16.99/9.62/2.12\% unsafe exposure; DINOv2 reaches 59.04/75.58/79.81\% and 31.28/12.24/4.62\%. Video-cluster sensitivity and person/non-person breakdowns are retained in the artifact registry.

\section{Paired end-to-end attribution tests}

\begin{table}[t]
\centering
\small
\setlength{\tabcolsep}{3.0pt}
\caption{Matched four-arm end-to-end results (\%). Every row uses the exact same cases and frozen decoder. Stress unsafe is unauthorized memory use on non-clean interventions (lower is better); positive recall prevents an abstention-only win. All-case unsafe, \nullmem{} specificity, and paired tests are reported in the supplement.}
\label{tab:e2e}
\resizebox{\linewidth}{!}{%
\begin{tabular}{lllrrrr}
\toprule
Source & Model & Metric & Raw & Anchor-triad & \method{} & \guard{} \\
\midrule
\multirow{3}{*}{RecordAuth-Diag} & Qwen3-VL-8B & Accuracy & 40.16 & 54.66 & 55.93 & 53.74 \\
 &  & Stress unsafe & 29.40 & 1.15 & 2.57 & 0.64 \\
 &  & Pos. recall & 37.05 & 39.88 & 40.34 & 35.08 \\
\multirow{3}{*}{RecordAuth-Diag} & Phi-3.5-Vision & Accuracy & 19.73 & 47.94 & 54.55 & 52.28 \\
 &  & Stress unsafe & 36.21 & 11.86 & 7.38 & 2.00 \\
 &  & Pos. recall & 5.30 & 44.87 & 46.65 & 38.75 \\
\multirow{3}{*}{RecordAuth-Diag} & CoViP & Accuracy & 40.41 & 59.51 & 60.70 & 57.97 \\
 &  & Stress unsafe & 32.28 & 2.07 & 3.39 & 0.85 \\
 &  & Pos. recall & 37.90 & 47.27 & 47.74 & 41.00 \\
\midrule
\multirow{3}{*}{DreamBooth} & Qwen3-VL-8B & Accuracy & 50.33 & 67.00 & 69.33 & 70.67 \\
 &  & Stress unsafe & 38.33 & 0.00 & 4.17 & 0.00 \\
 &  & Pos. recall & 57.62 & 58.57 & 60.95 & 58.10 \\
\multirow{3}{*}{DreamBooth} & Phi-3.5-Vision & Accuracy & 8.00 & 51.67 & 59.00 & 60.00 \\
 &  & Stress unsafe & 47.92 & 8.75 & 4.17 & 0.00 \\
 &  & Pos. recall & 2.86 & 54.76 & 57.62 & 55.71 \\
\multirow{3}{*}{DreamBooth} & CoViP & Accuracy & 51.00 & 73.33 & 76.33 & 77.67 \\
 &  & Stress unsafe & 39.58 & 2.50 & 5.00 & 0.83 \\
 &  & Pos. recall & 59.05 & 69.52 & 70.95 & 68.10 \\
\midrule
\multirow{3}{*}{DAVIS-2017} & Qwen3-VL-8B & Accuracy & 44.30 & 65.60 & 69.10 & 71.00 \\
 &  & Stress unsafe & 34.75 & 6.00 & 9.88 & 1.88 \\
 &  & Pos. recall & 44.00 & 61.29 & 65.57 & 60.86 \\
\multirow{3}{*}{DAVIS-2017} & Phi-3.5-Vision & Accuracy & 21.30 & 41.40 & 50.10 & 51.80 \\
 &  & Stress unsafe & 35.50 & 15.38 & 10.25 & 2.12 \\
 &  & Pos. recall & 7.86 & 38.29 & 42.43 & 39.14 \\
\bottomrule
\end{tabular}
}
\end{table}
\begin{table}[t]
\centering
\scriptsize
\setlength{\tabcolsep}{2.1pt}
\caption{Paired end-to-end attribution contrasts in percentage points. RecordAuth-Diag intervals resample query identities; DreamBooth resamples causal parents; DAVIS resamples tracks and additionally reports a case-weighted source-video sensitivity. McNemar $p$ is a case-level diagnostic, not clustered inference.}
\label{tab:e2e-paired}
\resizebox{\linewidth}{!}{%
\begin{tabular}{llllrllr}
\toprule
Source & Model & Contrast & Metric & $\Delta$ [95\% CI] & Cluster & Video CI & $p$ \\
\midrule
RecordAuth-Diag & Qwen3-VL-8B & Guard--Anchor & Accuracy & -0.92 [-2.49, +0.54] & identity & -- & 0.03 \\
 & Qwen3-VL-8B & Guard--Anchor & Stress unsafe & -0.51 [-0.86, -0.21] & identity & -- & 6.1e-05 \\
 & Qwen3-VL-8B & Guard--ATRA & Accuracy & -2.20 [-3.50, -1.05] & identity & -- & 7.44e-09 \\
 & Qwen3-VL-8B & Guard--ATRA & Stress unsafe & -1.93 [-2.65, -1.28] & identity & -- & 1.39e-17 \\
 & Phi-3.5-Vision & Guard--Anchor & Accuracy & +4.34 [+2.48, +6.37] & identity & -- & 3.77e-12 \\
 & Phi-3.5-Vision & Guard--Anchor & Stress unsafe & -9.86 [-13.00, -7.32] & identity & -- & 5.03e-88 \\
 & Phi-3.5-Vision & Guard--ATRA & Accuracy & -2.28 [-4.21, -0.66] & identity & -- & 4.38e-06 \\
 & Phi-3.5-Vision & Guard--ATRA & Stress unsafe & -5.39 [-6.58, -4.32] & identity & -- & 2.74e-48 \\
 & CoViP & Guard--Anchor & Accuracy & -1.54 [-3.51, +0.25] & identity & -- & 0.00131 \\
 & CoViP & Guard--Anchor & Stress unsafe & -1.22 [-2.14, -0.51] & identity & -- & 2.91e-11 \\
 & CoViP & Guard--ATRA & Accuracy & -2.74 [-4.33, -1.29] & identity & -- & 1.57e-10 \\
 & CoViP & Guard--ATRA & Stress unsafe & -2.54 [-3.42, -1.73] & identity & -- & 5.29e-23 \\
DreamBooth & Qwen3-VL-8B & Guard--Anchor & Accuracy & +3.67 [+1.67, +5.67] & parent & -- & 0.00342 \\
 & Qwen3-VL-8B & Guard--Anchor & Stress unsafe & +0.00 [+0.00, +0.00] & parent & -- & 1 \\
 & Qwen3-VL-8B & Guard--ATRA & Accuracy & +1.33 [-1.67, +4.33] & parent & -- & 0.454 \\
 & Qwen3-VL-8B & Guard--ATRA & Stress unsafe & -4.17 [-6.25, -2.08] & parent & -- & 0.00195 \\
 & Phi-3.5-Vision & Guard--Anchor & Accuracy & +8.33 [+4.33, +12.33] & parent & -- & 1.62e-06 \\
 & Phi-3.5-Vision & Guard--Anchor & Stress unsafe & -8.75 [-14.58, -3.75] & parent & -- & 9.54e-07 \\
 & Phi-3.5-Vision & Guard--ATRA & Accuracy & +1.00 [-1.67, +3.33] & parent & -- & 0.549 \\
 & Phi-3.5-Vision & Guard--ATRA & Stress unsafe & -4.17 [-6.25, -2.08] & parent & -- & 0.00195 \\
 & CoViP & Guard--Anchor & Accuracy & +4.33 [+1.33, +7.33] & parent & -- & 0.00443 \\
 & CoViP & Guard--Anchor & Stress unsafe & -1.67 [-5.00, +0.00] & parent & -- & 0.125 \\
 & CoViP & Guard--ATRA & Accuracy & +1.33 [-1.67, +4.33] & parent & -- & 0.454 \\
 & CoViP & Guard--ATRA & Stress unsafe & -4.17 [-6.25, -2.08] & parent & -- & 0.00195 \\
DAVIS-2017 & Qwen3-VL-8B & Guard--Anchor & Accuracy & +5.40 [+2.80, +8.00] & track & [+2.72, +8.02] & 1.91e-07 \\
 & Qwen3-VL-8B & Guard--Anchor & Stress unsafe & -4.12 [-6.38, -2.12] & track & [-6.31, -2.13] & 2.33e-10 \\
 & Qwen3-VL-8B & Guard--ATRA & Accuracy & +1.90 [-0.80, +4.30] & track & [-0.74, +4.34] & 0.0503 \\
 & Qwen3-VL-8B & Guard--ATRA & Stress unsafe & -8.00 [-10.38, -5.88] & track & [-12.04, -5.23] & 1.08e-19 \\
 & Phi-3.5-Vision & Guard--Anchor & Accuracy & +10.40 [+8.20, +12.60] & track & [+7.96, +12.90] & 2.87e-21 \\
 & Phi-3.5-Vision & Guard--Anchor & Stress unsafe & -13.25 [-16.50, -10.12] & track & [-16.88, -10.00] & 2.47e-32 \\
 & Phi-3.5-Vision & Guard--ATRA & Accuracy & +1.70 [-0.30, +3.50] & track & [-0.40, +3.60] & 0.043 \\
 & Phi-3.5-Vision & Guard--ATRA & Stress unsafe & -8.12 [-10.62, -6.00] & track & [-11.79, -5.29] & 5.42e-20 \\
\bottomrule
\end{tabular}
}
\end{table}

The track- and video-cluster intervals agree on every DAVIS safety contrast. The
guard's accuracy advantage over \method{} is not significant under either bootstrap,
whereas its unsafe-use reduction is. The DreamBooth Qwen anchor-triad equality and
CoViP uncertainty are retained in the same table rather than summarized as universal
success.

\section{Independent personalized-pipeline confirmation}
\label{app:rap-transfer}

RAP-Phi3-mini is independently authored and post-trained to consume retrieved
personal concepts. We pin checkpoint revision \texttt{0264b32f}, RAP source
revision \texttt{3c0c9c2e}, and OpenAI CLIP ViT-L/14-336 revision
\texttt{ce19dc91}. Each RecordAuth-Diag card becomes one RAP record: its image is the
retrieval key, while opaque subject, location, timestamp, and status form
\texttt{Name}/\texttt{Info}. The query image is first, followed by deterministic
full-frame cosine top-2 records in RAP's released multi-image syntax. Full-frame
retrieval replaces RAP's detector crops because the diagnostic views are already
single-subject images; we do not call this an exact deployment reproduction.

The adapter was debugged on 30 validation parents. An initial over-explicit refusal
prompt produced 100\% abstention and is retained as a failed run. The locked native
answering prompt was then evaluated once on test. Because one clean test parent had
already been inspected during an integration smoke test, its complete ten-condition
parent is excluded, leaving 3,680 cases, 368 parents, and 86 query identities. No
test-conditioned prompt, top-$k$, threshold, or parser change followed.

\begin{table}[t]
\centering
\scriptsize
\setlength{\tabcolsep}{2.3pt}
\caption{Independent RAP-Phi3 transfer (\%, 95\% query-identity-cluster CI; 3,680 cases, 368 causal parents). The frozen guard sees the same cases and can only filter cards before native CLIP top-2 retrieval.}
\label{tab:rap-transfer}
\resizebox{\linewidth}{!}{%
\begin{tabular}{lrrrrr}
\toprule
Arm & Clean recall & Stress unsafe & \nullmem{} specificity & All accuracy & Positive recall \\
\midrule
Raw & 27.45 [22.15,33.33] & 36.99 [34.87,39.51] & 0.00 [0.00,0.00] & 22.12 [19.78,24.68] & 31.60 [28.33,35.17] \\
Strict guard & 19.97 [15.08,25.07] & 1.56 [0.49,3.50] & 96.20 [93.48,98.05] & 47.58 [44.17,51.11] & 26.75 [22.31,31.61] \\
\midrule
Guard $-$ raw & $-7.47$ [$-10.71$,$-4.69$] & $-35.43$ [$-38.13$,$-32.84$] & $+96.20$ [93.56,98.06] & $+25.46$ [23.18,27.64] & -- \\
\bottomrule
\end{tabular}}
\end{table}
\begin{table}[t]
\centering
\scriptsize
\setlength{\tabcolsep}{3.0pt}
\caption{RAP generator attribution (\%, 95\% query-identity-cluster CI). The primary population is frozen before each context intervention.}
\label{tab:rap-attribution}
\resizebox{\linewidth}{!}{%
\begin{tabular}{llr}
\toprule
Population / intervention & Endpoint & Rate [CI] \\
\midrule
Raw unsafe ($n=1{,}105$), no memory & Same raw prediction & 0.00 [0.00,0.00] \\
Raw unsafe, nonce relabel & Exact displayed-nonce copy & 38.10 [34.06,42.80] \\
Raw unsafe, nonce relabel & Same-event exact tracking & 26.97 [23.29,30.97] \\
Raw unsafe, record shuffle & Same-event tracking & 66.61 [63.05,70.04] \\
Clean correct ($n=101$), nonce relabel & Exact displayed-nonce copy & 31.68 [23.26,40.45] \\
\bottomrule
\end{tabular}}
\end{table}

The guard's safety decrease is paired with a significant clean-recall cost. Exact
McNemar tests give $p<10^{-14}$ for clean recall and $p<10^{-294}$ for stress
unsafe use; identity-cluster intervals, not these case-level tests, remain primary.
Nonce and shuffle hold visual keys and their retrieval scores fixed while altering
record text; removal exposes no card. Thus nonce and same-event endpoints cannot be
explained by an unchanged visual match alone.

\section{Role-level split and dependence audit}
\label{app:role-audit}

The fold constructor assigns an identity to one split before any causal cases are
expanded. We nevertheless audit the serialized manifests, because a query-only audit
would miss leakage through distractor, anchor, or card roles. For every record we
enumerate query identity/image, every candidate-group identity, trusted-anchor image,
and memory-card image. Table~\ref{tab:role-audit} reports the resulting all-role sets.

\begin{table}[t]
\centering
\scriptsize
\setlength{\tabcolsep}{2.5pt}
\caption{Serialized all-role split audit. Counts are train/validation/test. Overlap
columns count identities or images appearing in more than one split within a fold;
``viol.'' counts records containing a role assigned outside their query split.}
\label{tab:role-audit}
\begin{tabular}{lrrrrr}
\toprule
Fold & Cases & Identities & Images & ID/image overlap & Viol. \\
\midrule
0 & 2320/700/670 & 53/16/18 & 755/226/218 & 0/0 & 0 \\
1 & 2300/670/720 & 51/18/18 & 746/218/235 & 0/0 & 0 \\
2 & 2190/720/780 & 51/18/18 & 711/235/253 & 0/0 & 0 \\
3 & 2090/780/820 & 52/18/17 & 679/253/267 & 0/0 & 0 \\
4 & 2170/820/700 & 54/17/16 & 706/267/226 & 0/0 & 0 \\
\bottomrule
\end{tabular}
\end{table}

Each fold contains the same 87 identities and 1,199 images overall; no group-role
identity is absent from the query-role assignment. Repeated use remains substantial
within a split (up to 528 group/anchor and 2,648 card occurrences per identity), so
query-identity resampling captures the main ownership cluster but not every shared-
distractor dependency. The audit therefore removes a leakage explanation without
turning the existing intervals into full learning-procedure uncertainty.

As a complementary diagnostic, we delete each identity together with every case in
which it appears as query, candidate group, anchor, or card, then recompute the
principal guard--anchor and guard--\method{} accuracy/unsafe contrasts. Across Qwen
and CoViP, all leave-one-identity-out estimates retain the full-data sign (zero sign
reversals). This tests leverage by repeated non-query roles; it is not a new split,
retraining run, or confidence interval over the fold-construction procedure.

\section{Frozen representation-to-policy analyses}
\label{app:policy-gap}

The public LSD diagnostic contains 50 identity clusters with ten contexts each. We
serialize four conditions per row: authorized original, gray-image missing mapping,
absent-persona query, and a derangement that preserves all real images and dialogues
but permutes their binding. Five fixed folds use 30/10/10 identities for train/
validation/test. Linear logistic probes select regularization and threshold on source
validation identities only. The primary feature is the final prompt-token state at
layer 24; controls combine null likelihood and image-similarity scalars. Split-crossing
absent-persona rows are excluded, leaving 1,936 cases for the all-invalid comparison.

\begin{table}[t]
\centering
\scriptsize
\setlength{\tabcolsep}{2.2pt}
\caption{Frozen representation audit. Within-checkpoint cells are out-of-fold AUROC;
brackets are identity-cluster 95\% intervals. Cross-checkpoint rows fit and calibrate
only on the source checkpoint; macro is mean$\pm$SD over target-fold AUROCs. BA uses
the source validation threshold on the target and exposes nonportable calibration.}
\label{tab:policy-gap-full}
\resizebox{\linewidth}{!}{%
\begin{tabular}{llrrrrr}
\toprule
Train/evaluate & Task & Hidden pooled [CI] & Scalar pooled [CI] & Target macro & Source-thr. BA & LOO mispair \\
\midrule
Base/Base & all invalid & .949 [.926,.968] & .802 [.770,.831] & -- & .872 & .599 \\
CoViP/CoViP & all invalid & .966 [.944,.983] & .810 [.788,.830] & -- & .908 & .646 \\
Base/CoViP & all invalid & .896 [.864,.931] & -- & .960$\pm$.019 & .686 & -- \\
CoViP/Base & all invalid & .876 [.834,.921] & -- & .948$\pm$.030 & .698 & -- \\
Base/CoViP & mispair & .871 [.839,.907] & -- & .958$\pm$.015 & .777 & -- \\
CoViP/Base & mispair & .820 [.780,.869] & -- & .924$\pm$.020 & .732 & -- \\
\bottomrule
\end{tabular}}
\end{table}

The within-checkpoint mispair AUROCs in Table~\ref{tab:policy-gap-full} are
.926/.964 for Base/CoViP, compared with .608/.603 for the scalar control. A layer
sweep rises from chance at layer 0 to its maximum near layer 24; paired hidden scores
rank original above derangement in 97.8/99.6\% of rows. Cross-checkpoint transfer is
meaningful because the checkpoints share architecture and aligned initialization, but
it is not cross-architecture evidence. The gap between pooled and fold-macro AUROC and
the moderate transferred BA show that ranking direction is more stable than global
calibration.

The simulated gate uses only out-of-fold readouts. It retains 82/83 and 115/121
correct originals while leaving 50 and 44 invalid false attributions (3.48/3.06\%).
This estimates recoverable policy headroom; it does not validate deployment under
shift. The leave-mispair-out column trains on missing and absent-persona negatives and
then evaluates a never-trained derangement. Its weak .599/.646 result is retained.
Finally, shifting the discriminative layer-24 direction by $-5$ and $-10$ in a
three-case prefill steering smoke test changes zero outputs. Decodability is neither a
mechanism-general axis nor a demonstrated causal control direction.

\section{Structured same-MLLM verifiers and reference robustness}
\label{app:staged-verifier}

The staged verifier is frozen before all RecordAuth-Diag inference. The group prompt presents
the query, every enrollment referent, every image-backed card, and active/missing/stale
status, but hides record locations and expected answers. It requires exactly one
\texttt{GROUP\_k} token or \texttt{NULL}. The optional edge prompt then presents only
the selected group's referent, query, and labelled cards and requires an exact
\texttt{KEEP: CARD\_i,...} list or \texttt{NULL}. Any parse failure maps to operational
\nullmem{}. Both prompts use greedy decoding and the frozen Qwen3-VL or CoViP checkpoint;
neither receives RecordAuth-Diag labels, demonstrations, training, or threshold selection.

The later one-pass control receives exactly the full-bank evidence of the group prompt
but must emit either \texttt{GROUP\_k; KEEP: CARD\_i,...} or \texttt{NULL} in one
call. Group and card indices must be in-range, unique, and sorted; every other output
fails closed. Its prompt, parser, two checkpoint revisions, and all 3,690 case IDs were
hash-frozen before output generation. It is an information-matched structured control,
not a compute-matched one: Qwen/CoViP use 11.25M input tokens and 3,583/3,517 verifier
seconds, versus 13.37/13.38M and 3,421/3,440 seconds for group plus edge. The joint
output is longer, so fewer input tokens do not imply lower measured latency.

Every decision file covers the same 3,690 test cases exactly. The unchanged natural
answer generator receives only the retained cards; a group/edge rejection invokes a
hard bypass without generation. For Qwen, full-bank routing uses 11.25M input tokens
and 2,294 seconds; the typed pass adds 2.11M tokens and 1,127 seconds. CoViP uses the
same 11.25M group tokens in 2,273 seconds and adds 2.13M edge tokens in 1,167 seconds.
Generation cost falls as more cases bypass; the archived structured-control summaries
retain the corresponding per-case totals. Peak verifier memory is 18.13~GiB on an
RTX~4090.

For the reference audit, each of 156 DAVIS tracks contributes two eligible temporal
views unused by the original anchor, cards, or query (312 assets), plus blurred and
center-occluded transforms of its original anchor (312 assets; 624 total). Exact and
perceptual-hash audits find no forbidden temporal-view overlap. The wrong view for
each group is the frozen-encoder nearest different track in the same candidate pool.
The 2-of-3 wrapper retains a group and each edge independently by majority vote; ties
and empty results return \nullmem{}. The quorum is applied to the original validation-
locked \method{} outputs without retraining or threshold changes. Primary intervals
resample tracks; a DAVIS-video sensitivity gives the same direction. With three clean
views, SigLIP2/DINOv2 reach 83.78/74.10\% accuracy and 9.55/11.92\% unsafe exposure,
close to the one-wrong quorum rows. The one-clean/two-wrong construction is retained
because the majority assumption predicts failure rather than repair.

\section{Omni-Persona visual protocol and complete routing results}
\label{app:omni}

We pin Omni-Persona code commit \texttt{818b0804} and dataset revision
\texttt{a5f99514}. All 231 visual items and 860 referenced assets pass path and hash
audits. Each item interleaves four persona contexts; its 115 answerable, 54 attribute-
absent, and 62 target-absent questions collapse to 172 unique source/query clusters.
Because the released visual item provides one context reference per persona, the same
reference is used as enrollment anchor and one-card group. This passes structural
mapping but removes anchor--card diversity: external \method{} performance is a hard
schema-shift test, not another RecordAuth-Diag edge-corruption result.

Raw and strict arms use the released context order and deterministic decoding. Frozen
SigLIP2 and \method{} use only non-Omni thresholds. The integrated verifier presents
the four context images and query image to the same frozen MLLM, requires exactly one
context index or \nullmem{}, and runs before the unchanged answer prompt. Table
\ref{tab:omni-routing} reports exact, judge-free routing.

\begin{table}[t]
\centering
\scriptsize
\setlength{\tabcolsep}{1.8pt}
\caption{Judge-free Omni routing with 95\% source-sample-cluster intervals (\%).
T-abs. is target-absent \nullmem{} specificity; A-abs. is correct identity routing
when the selected context lacks the queried attribute.}
\label{tab:omni-routing}
\resizebox{\linewidth}{!}{%
\begin{tabular}{lrrrrr}
\toprule
Router & Exact route [CI] & Unsafe [CI] & Pos. recall [CI] & T-abs. [CI] & A-abs. [CI] \\
\midrule
Frozen \method{} & 55.84 [47.64,63.95] & 3.03 [0.43,6.67] & 42.60 [33.54,51.76] & 91.94 [80.00,100] & 44.44 [30.77,58.14] \\
Qwen router & 90.91 [85.90,95.32] & 1.30 [0.00,3.51] & 89.35 [83.05,94.84] & 95.16 [87.04,100] & 90.74 [82.14,98.00] \\
CoViP router & 91.77 [86.96,95.95] & 3.90 [0.88,7.69] & 93.49 [88.76,97.48] & 87.10 [74.24,97.83] & 90.74 [82.14,98.04] \\
\bottomrule
\end{tabular}}
\end{table}

The integrated result is a strong architecture-equivalent baseline against the claim
that a separate learned gate is necessary. It supports the decision decomposition:
the frozen answer model can verify identity when forced to emit an explicit routing
action. It does not check whether the chosen biography states the requested attribute.
Frozen SigLIP2 similarity reaches 71.43\% exact routing and 10.82\% unsafe routing.
Order sensitivity remains nonzero. Four-order unanimous voting, added only after the
single-order outputs were seen, lowers unsafe routing to 0.87/1.30\% for Qwen/CoViP but
also lowers positive recall to 81.66/89.35\%; it is marked exploratory and is excluded
from confirmatory claims.

\begin{table}[t]
\centering
\scriptsize
\setlength{\tabcolsep}{3.2pt}
\caption{Why filtering is not an empty action (\%). ``Filtered'' invokes the released
answer-oriented generator with only the selected context; ``+NULL bypass'' returns a
fixed abstention without invoking it when routing rejects. Cal uses the local
Phi-3.5 sensitivity judge; target-absent false answer is deterministic.}
\label{tab:omni-bypass}
\begin{tabular}{llrrrr}
\toprule
Model & Verifier & Filtered Cal & Bypass Cal & Filtered false & Bypass false \\
\midrule
Qwen & frozen \method{} & 36.36 & 56.28 & 64.52 & 3.23 \\
Qwen & integrated & 47.19 & 64.07 & 64.52 & 3.23 \\
CoViP & frozen \method{} & 47.19 & 54.98 & 25.81 & 3.23 \\
CoViP & integrated & 54.55 & 60.61 & 29.03 & 6.45 \\
\bottomrule
\end{tabular}
\end{table}

The composer audits two invariants over all four bypass arms: every covered prediction
is byte-identical to the filtered run, and every rejected prediction is exactly the
fixed abstention. Thus the difference in Table~\ref{tab:omni-bypass} is caused by making
\nullmem{} operational, not regeneration or prompt editing. The benchmark's official
GPT-5.4-mini judge was unreachable from the execution environment; all free-form
metrics are explicitly labeled as local sensitivity measurements. No AI judge is
described as human validation.

\section{External all-negative transfer}

\begin{table}[t]
\centering
\scriptsize
\setlength{\tabcolsep}{2.4pt}
\caption{Actual ordinary-versus-authorized capability after an audited \nullmem{} decision (\%). Values are base$\rightarrow$authorized. Exact output agreement and exposed-card count audit the bypass rather than inferring preservation from a proxy.}
\label{tab:capability}
\begin{tabular}{llrrrrr}
\toprule
Task & Model & $n$ & Accuracy & Task-specific & Output agree. & Cards \\
\midrule
MMVP & Qwen3-VL-8B & 300 & 79.33$\rightarrow$79.33 & Strict-pair: 62.00$\rightarrow$62.00 & 100.00 & 0 \\
MMVP & Phi-3.5-Vision & 300 & 66.33$\rightarrow$66.33 & Strict-pair: 38.67$\rightarrow$38.67 & 100.00 & 0 \\
POPE & Qwen3-VL-8B & 9000 & 88.47$\rightarrow$88.47 & F1-yes: 87.75$\rightarrow$87.75 & 100.00 & 0 \\
POPE & Phi-3.5-Vision & 9000 & 79.92$\rightarrow$79.92 & F1-yes: 75.60$\rightarrow$75.60 & 100.00 & 0 \\
\bottomrule
\end{tabular}
\end{table}

On 500 POPE/COCO cases, SigLIP2 similarity/graph/anchor-triad/\method{}/guard produce 7/8/7/0/0 false authorizations. DINOv2 produces 104/196/78/2/1. On 300 MMVP cases, the corresponding SigLIP2 counts are 3/3/3/0/0 and DINOv2 counts are 35/97/41/1/1. These are gate decisions only. They motivate but do not substitute for the registered generator-level contamination tests.

\begin{table}[t]
\centering
\scriptsize
\setlength{\tabcolsep}{2.5pt}
\caption{Generator-level transfer on independent all-negative banks (\%). Cells are raw$\rightarrow$guard with guard-minus-raw parent-cluster 95\% intervals. ``Accuracy'' is task-defined \nullmem{} correctness rather than ordinary VQA accuracy; abstention is shown explicitly so refusal cannot masquerade as repair.}
\label{tab:external-e2e}
\resizebox{\linewidth}{!}{%
\begin{tabular}{llrrrr}
\toprule
Source & Model & $n$ & \nullmem{} acc. $\Delta$ [CI] & Unsafe $\Delta$ [CI] & Abstain \\
\midrule
MMVP & Qwen3-VL-8B & 300 & 100.00$\rightarrow$100.00 (+0.00 [+0.00,+0.00]) & 0.00$\rightarrow$0.00 (+0.00 [+0.00,+0.00]) & 100.00$\rightarrow$100.00 \\
MMVP & Phi-3.5-Vision & 300 & 100.00$\rightarrow$100.00 (+0.00 [+0.00,+0.00]) & 0.00$\rightarrow$0.00 (+0.00 [+0.00,+0.00]) & 100.00$\rightarrow$100.00 \\
POPE/COCO & Qwen3-VL-8B & 500 & 100.00$\rightarrow$100.00 (+0.00 [+0.00,+0.00]) & 0.00$\rightarrow$0.00 (+0.00 [+0.00,+0.00]) & 100.00$\rightarrow$100.00 \\
POPE/COCO & Phi-3.5-Vision & 500 & 100.00$\rightarrow$100.00 (+0.00 [+0.00,+0.00]) & 0.00$\rightarrow$0.00 (+0.00 [+0.00,+0.00]) & 100.00$\rightarrow$100.00 \\
\bottomrule
\end{tabular}
}
\end{table}

At generator level, both raw and guarded Qwen/Phi paths abstain on every custom
POPE and MMVP contamination case. This is a retained null result: the prompts do
not force either decoder to consume the irrelevant bank, so these sets demonstrate
neither a repair gain nor positive personalization. The ordinary VQA runs in
Table~\ref{tab:capability} answer the benchmark questions and serve the separate
capability-preservation claim.

\section{Failure retention and hardware audit}
\label{app:failure-audit}

The original strict-guard implementation intersected per-run decisions before majority voting. Because voting and intersection do not commute, the resulting ensemble could add an edge relative to the separately voted \method{} output. That output is archived and excluded. The corrected pipeline ensembles first and intersects once; audits across ten sources find zero added edges, index inconsistencies, or newly introduced unsafe cases. A registered 300-case rerun on A100 reproduces A800 outputs exactly for Qwen and Phi, raw and guarded: exact text, parsed predictions, answer flags, and unsafe flags all agree in 100\% of cases.

The DAVIS-56 cost registry counts actual non-bypassed generation calls. Guard-only
generation uses 285,435 input and 5,859 output tokens in 300.576 seconds; the raw path
adds 2,032,209 input and 5,485 output tokens in 538.978 seconds. Serialized dual-path
cost is therefore 2,317,644 input tokens, 11,344 output tokens, and 839.554 seconds.
The counterfactual policy and path-HGB consume these same saved candidates. Timed CPU replays with saved
SigLIP2 embeddings take 3.96 and 14.05 seconds, respectively, including process/model
loading but excluding image embedding; these are audit runtimes, not online latency.

\begin{table}[t]
\centering
\small
\caption{A800-to-A100 deterministic reproducibility (\%) on the registered 300-case subset.}
\label{tab:hardware}
\begin{tabular}{lrrrrr}
\toprule
Model / arm & $n$ & Text agree. & Parsed agree. & Accuracy & Unsafe \\
\midrule
Qwen / Raw & 300 & 100.00 & 100.00 & 39.33$\rightarrow$39.33 & 27.00$\rightarrow$27.00 \\
Qwen / Guard & 300 & 100.00 & 100.00 & 49.00$\rightarrow$49.00 & 1.67$\rightarrow$1.67 \\
Phi / Raw & 300 & 100.00 & 100.00 & 21.00$\rightarrow$21.00 & 22.67$\rightarrow$22.67 \\
Phi / Guard & 300 & 100.00 & 100.00 & 41.33$\rightarrow$41.33 & 5.67$\rightarrow$5.67 \\
\bottomrule
\end{tabular}
\end{table}

\section{Additional measurement and comparison limitations}
\label{app:additional-limits}

The CoViP authors describe their downstream LSD diagnostic images as evaluation-only~\citep{arxiv260203454},
but we cannot independently audit the checkpoint's complete training corpus; public
image overlap in foundation-model pretraining also remains possible. Frozen base Qwen
and Phi separate the CoViP-specific post-training effect, not generic pretraining
exposure. RAP supplies an independent checkpoint lineage, but its complete training
corpus and foundation-model image overlap are likewise not independently auditable.
Synthetic nonce labels and paired within-parent changes reduce the value of
memorizing a diagnostic answer, but do not make public-image provenance irrelevant.

Generator attribution is automatic and deliberately conservative: its headline
requires suffix-preserving equality with a synthetic location displayed after an
intervention, with the ordinary parser reported only as a sensitivity. This makes
nonce copying causally auditable but does not validate semantic attribution in
free-form personal narratives. A blinded 180-item packet has been specified for two
human annotators; until those judgments are actually collected, we make no claim of
human agreement or semantic-parser validity. No-memory and oracle-authorized runs are
diagnostic endpoints, not usable deployment policies.

Flat-LogReg's deficit establishes only that a linear fixed aggregate is insufficient.
The 55,773-parameter Flat-MLP closes the capacity confound: it statistically matches
\method{} on SigLIP2 accuracy and unsafe exposure, but is significantly worse on both
under DINOv2. Learned hierarchical aggregation is therefore an encoder-dependent
robustness choice, not a demonstrated universal necessity or optimal architecture.

\section{Reproducibility contract}
\label{app:reproducibility}

The complete code and evidence package is available through the
\artifactlink.

The immutable V3 package is generated only if a fail-closed registry verifies all
52 original inference endpoints, 21 analyses, and 14 static evidence summaries.
A separate post-review registry requires all 12 core model--intervention endpoints
plus five RAP arms. Core arms have 3,690 unique cases each (44,280 aligned rows);
RAP arms have 3,680 each (18,400 rows) after the registered parent exclusion. Both
require the frozen
10,000-replicate analysis, a current ten-test regression log, and an explicitly
pending rather than completed human-annotation packet. Each endpoint records expected
case count, unique case key, model name, source-manifest hash, output hash, and
authorizer hash. Partial JSONL files are never opened by the paper exporter. The
package includes source revisions and licenses, manifests, frozen embeddings, 25
checkpoints per encoder, validation grids, ensemble decisions, raw generator outputs,
logs, paired bootstrap units, environment versions, and retained failures. A
table-level manifest maps every paper-facing row to canonical analysis JSON and
SHA-256. The release is built under a non-final staging name, recompiles
its archived source in isolation, checks normalized PDF-text identity, and is moved
to its final path only after numerical, privacy, font, and citation audits pass.
An additive expansion registry separately requires four complete 3,690-case
anchor-visible runs (14,760 rows), 50 capacity-control checkpoints, three clean
analysis audits, and three source-hashed tables; it does not mutate the frozen
V3 or post-review registries.

The frozen revisions are Qwen3-VL-8B-Instruct
\texttt{0c351dd0}, CoViP-Qwen3-VL-8B-GSPO \texttt{2283a306},
Phi-3.5-Vision \texttt{12b77fb4}, Gemma-3-4B-IT \texttt{4b66b5f8},
SigLIP2-base-patch16-224
\texttt{75de2d55}, and DINOv2-base \texttt{f9e44c81}; full 40-character
revisions and metadata hashes are stored in the registry. RAP-Phi3-mini, RAP source,
and its CLIP retriever are pinned to \texttt{0264b32f}, \texttt{3c0c9c2e}, and
\texttt{ce19dc91}; its five-arm audit independently verifies case alignment and
paper-facing integer counts. Qwen/CoViP and Gemma use
3,136--100,352 pixels per image; Phi uses one crop with SDPA. Core decoding is greedy
with at most 48 new tokens; RAP uses 32. Every card is serialized with group/card index, opaque
subject label, timestamp, active/stale status, location, and its image (or an explicit
missing-image marker), followed by the query image and the one-line boxed-answer
instruction. The saved local and Phi environments include complete package freezes,
CUDA/GPU reports, commands, and model-config hashes.

The Round-1 revision manifest is additive and leaves those frozen registries intact.
It hashes the serialized all-role split audit; both checkpoints' frozen LSD feature,
probe, policy-gap, and cross-checkpoint summaries; Omni-Persona commit/dataset
revisions, 231-row predictions, 860-asset download audit, exact routing decisions,
NULL-bypass invariants, local-judge outputs, analyses, scripts, commands, and failed/
recovered logs. It distinguishes confirmatory single-order routing from the post-output
permutation ensemble and records failed access to the official judge, preventing local
sensitivity scores from being silently promoted to official results.

The Round-2 additive registry freezes the staged group/edge prompts, exact parsers,
five per-fold decision sets, and audited 3,690-case aggregates for Qwen and CoViP. It
separately hashes four 3,690-case generator aggregates, clustered multi-arm analyses,
token/time accounting, and bypass counts. The reference-corruption registry contains
156 DAVIS tracks, 1,560 cases, 312 nonoverlapping temporal views, 312 controlled
anchor transforms, both frozen encoders,
all single/quorum decisions, track/video analyses, and two 1,000-case Qwen generator
arms. A fail-closed table builder requires 14 frozen summaries and hashes every
generated snippet; this file inlines their numeric content. The all-231-item judge-free
Omni audit remains separate from local-judge sensitivity files.

The Round-3 additive registry freezes the one-pass joint prompt and parser, all ten
fold/model output shards, two exact 3,690-case aggregate audits, clustered gate
contrasts, two hard-bypass generator runs, and paired end-to-end analyses. It also
hashes the four-interface condition table and the failed/recovered CoViP fold log; the
failed attempt produced no cases and the clean retry covers the frozen fold exactly.
The main-paper predicate decomposition is generated directly from the locked equal-budget
summary by \texttt{scripts/make\_predicate\_decomposition\_figure.py}; its sidecar records
the input hash, field path, and every plotted or annotated count. The separate operating-point
figure is generated without refitting by
\texttt{scripts/make\_vmm\_main\_analysis\_figure.py}; its sidecar records the active
manuscript hash and plotted table rows. The derived same-union ATRA control is
replayed by \texttt{scripts/evaluate\_dual\_candidate\_atra\_filter.py} with zero new
MLLM calls; its input hashes and paired bootstrap are recorded in the artifact registry.
The group-relevance controls are replayed by
\texttt{scripts/analyze\_relevance\_baseline.py}; the locked summary records development-only
threshold selection, score recovery and fail-closed counts, token-estimation status, and
the predicate-level 27/4/2/0 unsafe reduction.
An additive Gemma audit hashes the raw full-bank arm, the group/\nullmem{} and typed-edge
pipeline arms, the unchanged 48-token greedy decoding contract, the frozen format-tolerant
parser, all 3,690 parsed outputs per arm, and the 10,000-replicate identity-cluster analyses.

\end{document}